\documentclass{article}

\usepackage[nonatbib,preprint]{neurips_2023}

\usepackage[utf8]{inputenc} 
\usepackage[T1]{fontenc}    
\usepackage{hyperref}       
\usepackage{url}            
\usepackage{booktabs}       
\usepackage{amsfonts}       
\usepackage{nicefrac}       
\usepackage{microtype}      
\usepackage{xcolor}         
\usepackage{graphicx}
\usepackage{amsmath,amssymb,amsthm}
\newtheorem{prop}{Proposition}
\newtheorem{theorem}{Theorem}
\newtheorem{corollary}{Corollary}
\usepackage{relsize}
\usepackage[backend=bibtex]{biblatex}
\bibstyle{abbrv}

\title{Double-descent curves in neural networks: \\
           a new perspective using Gaussian processes}

\author{Ouns El Harzli \\
Department of Computer Science\\
  University of Oxford\\
  \texttt{ouns.elharzli@new.ox.ac.uk} \\
  \And Bernardo Cuenca Grau \\
  Department of Computer Science\\
  University of Oxford\\
  \And Guillermo Valle-Pérez \\
  Rudolf Peierls Centre for Theoretical Physics\\ 
    University of Oxford \\
    \And Ard A. Louis \\
  Rudolf Peierls Centre for Theoretical Physics\\ 
    University of Oxford \\
    }

\begin{document}

\maketitle

\begin{abstract}
Double-descent curves in neural networks describe the phenomenon that the generalisation error initially descends with increasing parameters, then grows after reaching an optimal number of parameters which is less than the number of data points, but then descends again in the overparameterized regime. In this paper, we use techniques from random matrix theory to characterize the spectral distribution of the empirical feature covariance matrix as a width-dependent perturbation of the spectrum of the neural network Gaussian process (NNGP) kernel, thus establishing a novel connection between the NNGP literature and the random matrix theory literature in the context of neural networks. Our analytical expression allows us to study the generalisation behavior of the corresponding kernel and GP regression, and provides a new interpretation of the double-descent phenomenon, namely as governed by the discrepancy between the width-dependent empirical kernel and the width-independent NNGP kernel.
\end{abstract}

\section{Introduction}
\label{submission}

Deep learning has experienced unprecedented
success in a wide range of tasks \cite{deeplearning1,deep2,lecun2015deep,schmidhuber2015deep,deeplearning3}. Surprisingly, however, the best-performing Deep Neural Networks (DNNs)  operate in a highly over-parametrised regime, i.e.\ when the number of parameters in the model is  much larger than the number of training examples \cite{vgg}; this goes against conventional statistical wisdom of bias-variance trade-off which predicts that, in order to
avoid overfitting and obtain the best possible
generalisation, the number of parameters should be lower than the number of training examples \cite{rademacher, vcdim,shalev2014understanding}. 

The generalisation error of DNNs as a function of the number of parameters in the model
has been studied empirically ~\cite{belkin2019,Nakkiran2021dd}, and researchers have observed that it follows a double-descent curve
instead of the  classical U-shaped curve characteristic of the bias-variance trade-off. Specifically, for a fixed number of training examples, the generalisation error increases as the number of parameters approaches the number of training examples and, 
past this so-called \emph{interpolation threshold}, it starts decreasing again finding its global minimum when the number of parameters goes to infinity.  
Understanding these surprising observations
at a more fundamental level is an important
step towards tackling the deeper question as to \emph{why} DNNs generalise so well in practice~\cite{geiger2020perspective}. 


A number of mathematical frameworks for explaining the double-descent phenomenon in a variety of DNN architectures have been proposed in the literature  \cite{randomfeature,surprises,highdim,kernelreghighdim}. Assuming a teacher-student setting \cite{sompolinsky}, these
works derive an analytical expression for the generalisation error as a function of
the ratios $\gamma = \frac{n}{N}$ between the number $n$ of training examples and the width $N$ of the neural network and $\psi = \frac{n}{d}$ between the number of training examples and the dimension $d$ of the input. For a fixed value of $\psi$ and varying $\gamma$,
the generalisation error derived in these approaches follows 
a double descent curve; furthermore, the value of the generalisation error for a given  $\gamma$ and $\psi$ is obtained as a limit where $n$, 
$N$ and $d$ go to infinity while $\gamma$ and $\psi$ remain constant. 

A (largely orthogonal) line of research has studied the equivalence between infinitely-wide neural networks with random weights and Gaussian processes with a particular covariance function \cite{nngp}, typically referred to as the \emph{NNGP (or conjugate) kernel}. More precisely, in the limit of infinite width, the class of functions obtained by choosing the weights of the network at random
converges in distribution to a Gaussian process whose covariance function has a particular structure that can be defined inductively on the number of layers in the network. 

In this paper, we establish a novel connection between both of these orthogonal lines of research. Given a fully-connected neural architecture defined by the input dimension, the width of each layer, the activation function, and the distribution of the random weights, we derive an analytical expression for a \emph{width-dependent NNGP kernel} which generalises the so-called \emph{empirical covariance matrices} by \cite{randomfeature} to a kernel function.
We then exploit elements of Random Matrix Theory \cite{randommatrix, Livan_2018} to analytically compute, given a set of training examples, the spectral distribution of (the covariance matrix associated to) our width-dependent NNGP kernel. Although similar expressions have been computed in prior work \cite{spectrack}, ours is unique in that it is given as a function of the spectral distribution of the NNGP kernel; as
a result, our expression enables a new interpretation of the width-dependent spectral distribution as a perturbation of the width-independent spectral distribution that decreases as $\gamma$ tends to zero so that, in the limit, the width-dependent spectral distribution converges to that of an NNGP.
The kernel function and the the analytical formula for the spectral distribution that we propose allow us compute the generalisation error of both GP and kernel regression as a function of $\gamma$ and $\psi$. Similarly to \cite{randomfeature,surprises,highdim,specdecntk,specbiasandtaskalign} the value of the generalisation error at $\gamma$ and $\psi$ is computed as a limit
where $n$, 
$N$ and $d$ go to infinity while the ratios $\gamma$ and $\psi$ remain constant. Furthermore, for a fixed $\psi$ and varying $\gamma$, the generalisation error exhibits double descent behaviour.

Our approach requires only mild
assumptions on the network architecture and the data generating process. In particular, we require that the target function to be learnt has a bounded second moment with respect to the data distribution and we only require mild regularity assumptions (measurability and Lipschitzianity) for the nonlinear activation function in the network which  are satisfied by all commonly-used activations functions.

 Our results provide a new interpretation to the double-descent phenomenon where the behaviour of the generalisation error is governed by the discrepancy between the width-dependent empirical kernel characterising the network's architecture and the width-independent NNGP kernel of the limit Gaussian process.

\section{Preliminaries}

In this section we introduce the basic concepts underpinning our technical results. Throughout the paper, we will denote matrices by bold uppercase letters and vectors by bold lowercase letters.

\subsection{Elements of random matrix theory}

Random matrix theory \cite{randommatrix,Livan_2018} is the study of the spectral distributions of large random matrices (i.e.,  matrices whose elements are random variables).
The spectral measure $F_n$ of a given matrix with eigenvalues $\lambda_i$ is a measure over $x \in \mathbb{R}$ given by  $F_n(x) := \frac{1}{n} \sum_{i=1}^n \delta_{\lambda_i} (x)$, 
where $\delta_{\lambda_i} (x)$ is the Dirac measure at an eigenvalue $\lambda_i$.
When the matrix is random, the spectral measure becomes a random measure, often referred to as the empirical spectral distribution.

We study weak convergences (convergences in distribution) of spectral measures to nonrandom measures \cite{geronimo}. 
A  useful tool to manipulate spectral measures is the Stieltjes transform; for a measure $F$ supported on the real interval I, the Stieltjes transform is given as follows for each $z \in \mathbb{C}-I$:
$S_F (z) = \int_I{\frac{d F(\lambda)}{\lambda - z}}$.
There is a one-to-one correspondence between measures and their Stieltjes transforms, as per the inversion formula \cite{geronimo}: $F(x) = \lim_{y \to 0^+} S_F (x + i y)$ for $x \in I \setminus \{0\}$.
A sufficient condition for weak convergence of measures is to have pointwise convergence in their Stieltjes transforms \cite{geronimo}. 

We will rely on a famous result in random matrix theory. Consider $\mathbf{X} \in \mathbb{R}^{N \times n}$, a random matrix with i.i.d.\ entries drawn from $\mathcal{N} (0, \frac{1}{N})$ and $\Psi$ a nonrandom positive semi-definite matrix. Suppose that $\Psi$ has a limiting spectral measure $\mu$, and let $n, N \to \infty$ with fixed ratio $\gamma := \frac{n}{N}$, then the random matrix $\mathbf{\Psi}^{1/2} \mathbf{X}^T \mathbf{X} \mathbf{\Psi}^{1/2}$ has a limiting nonrandom spectral measure $\rho_\gamma^{MP} \boxtimes \mu$. The measure $\rho_\gamma^{MP} \boxtimes \mu$ is defined by its Stieltjes transform $S$, which solves the Marchenko-Pastur fixed-point equation \cite{marchenkopastur}:
\begin{equation}\label{marchenkopasturfixedpoint}
    S (z) = \int \frac{1}{x\left(1- \gamma - \gamma z S (z)\right) - z} d\mu (x).
\end{equation}
The measure $\rho_{MP}^{\gamma} \boxtimes \mu$ is called the Marchenko-Pastur map of $\mu$. In the particular case $\mathbf{\Psi} = \mathbf{I}_n$, $\mu$ is the Dirac measure at 1, and one recovers the Marchenko-Pastur distribution $\rho_{MP}^{\gamma}$.

\subsection{Neural network Gaussian processes}

A Gaussian process $f$ over a space $\mathbb{R}^d$ is a random scalar field
such that its evaluation at any collection of finitely many points $(f(x_1), ..., f(x_n))$ follows a multivariate Gaussian distribution. 

A Gaussian process is determined by a mean function $\mu: \mathbb{R}^d \mapsto \mathbb{R}$, and a covariance function $K: \mathbb{R}^d \times \mathbb{R}^d \mapsto \mathbb{R}$, which describe respectively the mean of the Gaussian distribution at each point and the covariance between the Gaussians at any two points. For Gaussian processes, the covariance function is a kernel, i.e. a positive semi-definite symmetric function \cite{gpforml}. We note $f \sim \mathcal{GP} (\mu, K)$.

We consider a random fully-connected neural network (FCN) with zero bias as in \cite{randomfeature}):
\begin{equation} \label{random nn}
    \mathbf{x}^l := \phi(\mathbf{h}^l)\qquad \mathbf{h}^l:= \mathbf{W}^l \mathbf{x}^{l-1} \qquad \forall l \in 1,...,L
\end{equation} 
where $N_0 := d$ is the dimension of the input space, $x^0 \in \mathbb{R}^d$ is an arbitrary input, $N_l$ is the width of the $l$-th layer, $\mathbf{h}^l := \mathbf{h}^l(\mathbf{x_0})$ is the preactivation of the $l$-th layer, the weight matrices $\mathbf{W}^l \in \mathbb{R}^{N_l \times N_{l-1}}$ have entries drawn i.i.d.\ from a Gaussian distribution $\mathcal{N} (0, \frac{1}{N_{l-1}})$, and $\phi$ is an arbitrary non-linear activation function acting componentwise. 

Applying successively the central limit theorem to each layer, the infinite-width limit of \eqref{random nn} yields a Gaussian process, called  the \emph{Neural Network Gaussian Process (NNGP)}. More precisely, if we let $N_0, ..., N_{L-1} \to \infty$, the $\mathbf{h}_i^L \sim \mathcal{GP} (\mu^L, K^L)$ are independent and defined 
inductively by layers as follows for all $\mathbf{x}, \mathbf{x}' \in \mathbb{R}^d$ and each $l \in 1,...,L$:
\begin{equation}\label{linearkernel}
    \mu^l (\mathbf{x}) = 0  \qquad  K^{\phi,0} (\mathbf{x}, \mathbf{x}') =  \mathbf{x}^T\mathbf{x}' 
\end{equation}
\begin{equation}
    \mathbf{h}_i^{l-1} \sim \mathcal{GP} (\mu^{l-1}, K^{\phi, l-1}) \qquad   K^{\phi,l}(\mathbf{x},\mathbf{x}')  =   \mathbb{E}_h \left( \phi(\mathbf{h}_i^{l-1} (\mathbf{x})) \phi(\mathbf{h}_i^{l-1} (\mathbf{x}'))\right) 
\end{equation}
The covariance function $K^{\phi,L}$ is called the \emph{NNGP kernel} or conjugate kernel \cite{daniely2}, which is determined by the
network depth $L$ and the activation function $\phi$; when these are clear from the context, we will simply denote it as $K$. There is potential for many subtleties in the way the infinite-width limits are approached \cite{matthews2018gaussian}, and we follow the approach in \cite{nngp} where infinite limits taken sequentially. 
The recursive formulae for the NNGP kernel have also been determined by \cite{poole} in the context of mean-field theory of random neural networks. We will use some of their techniques in our proofs.

The equivalence between randomly-initialised neural networks with infinite width and Gaussian processes with particular covariance functions is a general result which has been first established in \cite{neal} and revisited recently in \cite{nngp,matthews2018gaussian,novak2018bayesian,garriga2018deep,yang2019tensor,lee2020finite} and others. 

\subsection{Problem setup}

We consider the following teacher-student setting:
\begin{equation}\label{teacher}
    \mathbf{x} \sim \mathbb{P}_{d} \qquad \tau \sim \mathcal{N}(0,{\sigma_\tau}^2) \qquad y = f_d(\mathbf{x}) + \tau
\end{equation}
where $d$ is the dimension of the input space, $\mathbf{x} \in \mathbb{R}^d$ is the input feature vector, $\tau$ is the noise, $y$ is the output value, $\mathbb{P}_{d}$ is a family of probability distributions over $\mathbb{R}^d$ such that $\mathbb{E}_{\mathbf{x} \sim \mathbb{P}_d} (\mathbf{x}^T \mathbf{x})$ is bounded as $d \to \infty$; and $f_d: \mathbb{R}^d \longrightarrow \mathbb{R}$  is a family of functions, which verify that $\mathbb{E}_{\mathbf{x} \sim \mathbb{P}_{d}}  \left(f_d(\mathbf{x})^2\right)$ is bounded as $d \to \infty$. These assumptions, also taken in \cite{specdecntk,specbiasandtaskalign}, are quite mild as they only exclude pathological behaviors where the variances of the input or the output explode at infinity. 


We sample a number $n$ of examples, i.i.d.\ from the teacher. The training set then consists of:
\begin{equation} \label{training examples}
    \mathbf{X} = (\mathbf{x}_1,..., \mathbf{x}_n)^T \in \mathbb{R}^{n \times d} \qquad    \mathbf{y} = f_d(\mathbf{X}) + \mathbf{t} = (y_1,...,y_n)^T \in \mathbb{R}^n
\end{equation}
where $f_d(\mathbf{X}) = (f_d(\mathbf{x}_1), ..., f_d(\mathbf{x}_n))^T \in \mathbb{R}^n$ and $\mathbf{t} = (\tau_1, ..., \tau_n)^T \in \mathbb{R}^n$ is a noise sample.

We  study the behavior of the generalisation error of particular kernel regressions and Gaussian process regressions trained on this data.

\subsection{Gaussian process and kernel regression}
Consider an NNGP $z \sim \mathcal{GP} (0, K)$, where $K$ is the NNGP kernel obtained with the infinite-width limit of equation (\ref{random nn}), and taking the width of the output layer to be $1$, thus yielding an output $z$ in  $\mathbb{R}$. The standard Bayesian inference in Gaussian process regression \cite{gpforml} gives us that the prediction $\bar{z}$ conditionally on $(\mathbf{X}, \mathbf{y}, \mathbf{x})$ follows a Gaussian distribution $\bar{z} |\mathbf{X},\mathbf{y},\mathbf{x} \sim \mathcal{N} (\bar{\mu}, \bar{K})$ where:
\begin{equation}\label{nngp mean}
    \bar{\mu}_{\mathbf{X},\mathbf{y},\mathbf{x}} = \mathbf{k}_{\mathbf{x}, \mathbf{X}}^T (\mathbf{K}_{\mathbf{X},\mathbf{X}} + \sigma_\epsilon^2 \mathbf{I}_n)^{-1}\mathbf{y}  
\end{equation}
\begin{equation}\label{nngp variance}
    \bar{K}_{\mathbf{X},\mathbf{y},\mathbf{x}} = K(\mathbf{x},\mathbf{x}) - \mathbf{k}_{\mathbf{x}, \mathbf{X}}^T (\mathbf{K}_{\mathbf{X},\mathbf{X}} +  \sigma_\epsilon^2 \mathbf{I}_n)^{-1} \mathbf{k}_{\mathbf{x},\mathbf{X}}  
\end{equation}
if we assume a noise model $z|\mathbf{y} \sim \mathcal{N} (z, \sigma_\epsilon^2)$, and where $\mathbf{K}_{\mathbf{X},\mathbf{X}} \in \mathbb{R}^{n \times n}$ with $(\mathbf{K}_{\mathbf{X},\mathbf{X}})_{i,j} := K(\mathbf{x}_i,\mathbf{x}_j)$ and $\mathbf{k}_{\mathbf{x},\mathbf{X}} \in \mathbb{R}^n$ with $(\mathbf{k}_{\mathbf{x},\mathbf{X}})_i := K(\mathbf{x},\mathbf{x}_i)$. 

Furthermore, the mean prediction $\bar{\mu}_{\mathbf{X},\mathbf{y},\mathbf{x}}$ of GP regression is also the solution of kernel ridge regression with the same kernel $K$ and ridge parameter $\sigma_\epsilon^2$ \cite{gpforml}.

\section{A Gaussian process perspective on the double descent phenomenon}

We propose to exploit NNGPs to recover the double descent behavior. Our idea was motivated by the fact that, in previous studies of double descent in neural networks using random matrix theory \cite{randomfeature,surprises, highdim,kernelreghighdim}, the network width $N$ is indeed taken to infinity (along with the dimension of the input space, and the number of training examples). 



Our first step will be to characterise a counterpart of the NNGP kernel with finite width since a well-defined kernel is a prerequisite  to leverage the theories of GP regression and kernel regression. We were able to identify the random kernel underpinning the empirical covariance matrix of features by assuming that the width of the last layer is finite; as we will see, this already allows us to derive insights on the double-descent phenomenon.   

Then, we study the limiting spectral distribution of the empirical covariance matrix of features. We derive a non-trivial relationship between the spectral distributions of the empirical NNGP kernel random matrix and the actual NNGP kernel random matrix. 

Finally, we isolate the dependency to these spectral distributions in the generalisation errors of GP regression and kernel regression. This allows us to interpret the double-descent phenomenon as simply arising from the fact that the spectrum of the corresponding random kernel is a perturbation of that of the NNGP kernel limit whose prevalence varies as a function of $\gamma$, as reflected in the spectral distributions.

\subsection{A width-dependent random kernel}

Consider $z \sim \mathcal{GP} (0, K)$ obtained with the infinite width limit of (\ref{random nn}).  Finding a dependence with the width $N$ is not straightforward in the case of NNGPs because, at this point, the network width has already been taken to infinity. Our idea is therefore to study the behavior of a counterpart of the Gaussian process $z$ \emph{before} the width is taken to infinity. 

We denote  as $h^{L,N}$ the output of a random neural network defined as in (\ref{random nn}) with $L \geq 2$, $N_1, ..., N_{L-2} = \infty$, $N_{L-1} = N$ and $N_L = 1$, i.e.\ where all the widths have been taken to infinity (following \cite{nngp}) with the exception of the last one:
\begin{equation}
    h^{L,N} (\mathbf{x}_i) = \sum_{k =1}^N \mathbf{W}_k^L \phi \left( \mathbf{h}_k^{L-1} (\mathbf{x}_i) \right).
\end{equation}

\begin{prop}\label{unbiased estimator}
      The covariance matrix of the evaluations of $h^{L,N}$, conditional on the pre-activations, satisfies, for all pairs of training data points $\mathbf{x}_i, \mathbf{x}_j \in \mathbb{R}^d$ (rows of $\mathbf{X}$), that:
\begin{equation*}\label{kernel random matrix}
    \left(\mathbf{K}_{\mathbf{X}, \phi, \mathbf{h}^{L-1}}^N\right)_{i,j} := \mathbb{E}_{\mathbf{W}^L} \left( h^{L,N} (\mathbf{x}_i) h^{L,N} (\mathbf{x}_j) | \mathbf{h}_1^{L-1},..., \mathbf{h}_N^{L-1} \right) = \frac{1}{N} \sum_{k = 1}^N \phi \left( \mathbf{h}_k^{L-1} (\mathbf{x}_i) \right) \phi \left( \mathbf{h}_k^{L-1} (\mathbf{x}_j) \right)
\end{equation*} 
 where the expectation is thus taken over the last-layer weights $\mathbf{W}^L$, and it is an unbiased estimator of $K(\mathbf{x}_i, \mathbf{x}_j)$ with variance $\mathrm{Var} \left( \left(\mathbf{K}_{\mathbf{X}, \phi,\mathbf{h}^{L-1}}^N\right)_{i,j} \right) = \mathcal{O}_{N \to \infty} (\frac{1}{N})$ in standard big O notation.
\end{prop}

The random matrix $\mathbf{K}_{\mathbf{X}, \phi, \mathbf{h}^{L-1}}^N$ is the empirical covariance matrix of the features created by the NNGP (\ref{random nn}) before the last width is taken to infinity.

Conditionally on $\mathbf{X}$, the values $K (\mathbf{x}_i, \mathbf{x}_j)$ are constant and the $\left(\mathbf{K}_{\mathbf{X}, \phi, \mathbf{h}^{L-1}}^N\right)_{i,j}$ are random variables whose randomness stems from $
\mathbf{h}^{L-1}$. In turn, $\mathbf{K}_{\mathbf{X}, \phi,\mathbf{h}^{L-1}}^N$ satisfies the kernel property \cite{gpforml}:
\begin{equation} \label{kernel property}
    \sum_{i=1}^n \sum_{j=1}^n a_i a_j  \left(\mathbf{K}_{\mathbf{X}, \phi, \mathbf{h}^{L-1}}^N\right)_{i,j} = \frac{1}{N} \mathbf{a}^T \mathbf{\Phi}^T \mathbf{\Phi} \mathbf{a} \geq 0 \qquad  \forall \mathbf{a} = (a_1, ..., a_n)^T \in \mathbb{R}^n,
\end{equation}
where $\mathbf{\Phi} \in \mathbb{R}^{N \times n}$ with $\mathbf{\Phi}_{jk}= \phi \left(\mathbf{h}_j^{L-1} (\mathbf{x}_k) \right)$, and which holds for any realisation of the random matrix $\mathbf{\Phi}$. 
Note, however, that $\mathbf{K}_{\mathbf{X}, \phi, \mathbf{h}^{L-1}}^N$ does not technically define a kernel as it is not a well-defined function of $(\mathbf{x}, \mathbf{x}')$ but merely a countable family of random variables that can be indexed on $(\mathbf{x},\mathbf{x}')$. This is problematic in our setting since the covariance function in a Gaussian process
must be a kernel with respect to the full, continuous, space.

We next propose a way of
converting the aforementioned family of random variables into a random kernel, i.e.\ a kernel-valued random variable. 

\begin{theorem}\label{random field of kernels}
For $N \in (1, \infty)$, and $L \geq 2$ there exists a probability space  $(\Omega_N, \mathcal{A}_N, \mathbb{P}_N)$ and a random variable $K^{\phi, L,N} :  \Omega_{N} \longrightarrow \mathbb{R}^{(\mathbb{R}^d)^2}$ with image in the functional space $\mathbb{R}^{(\mathbb{R}^d)^2}$
such that :
\begin{enumerate}
    \item $K^{\phi, L,N} (\omega)$ is a kernel for all $\omega \in \Omega_N$,
    \item for all sets of points $\mathbf{X} = (\mathbf{x}_1, ..., \mathbf{x}_n)^T \in \mathbb{R}^{n \times d}$, the random matrix $\mathbf{K}_{\mathbf{X}, \phi, \mathbf{h}^{L-1}}^N$, and the random matrix $\mathbf{K}_{\mathbf{X},\mathbf{X}}^{\phi, L,N}$, defined as,
    \begin{equation}
        \mathbf{K}_{\mathbf{X},\mathbf{X}}^{\phi,L,N} : \Omega_N \longrightarrow \mathbb{R}^{n \times n}   \qquad \omega \mapsto \left(K^{\phi, L,N} ( \omega) (\mathbf{x}_i, \mathbf{x}_j) \right)_{ij \in [n]}
\end{equation}
follow the same distribution. In particular, $\mathbb{E}_{K^{\phi,L,N}} \left(K^{\phi,L,N} (\mathbf{x}_i, \mathbf{x}_j) \right)= K^{\phi, L} (\mathbf{x}_i, \mathbf{x}_j)$ for all $\mathbf{x}_i, \mathbf{x}_j \in \mathbb{R}^d$, where the expectation is taken over the random kernel function $K^{\phi,L,N}$.
\end{enumerate} 
\end{theorem}

We have thus defined a random variable $K^{\phi,L,N}$ over a functional space, whose realisations are kernel functions interpolating the random matrices of interest. When there is no ambiguity, we use $K^N$ to denote $K^{\phi,L,N}$. 
We can now study the random matrices $\mathbf{K}_{\mathbf{X},\mathbf{X}}^N$, whose randomness stems from the random kernel function $K^{N}$ and the random matrix $\mathbf{X}$, using the more convenient definition of $\mathbf{K}_{\mathbf{X}, \phi, \mathbf{h}^{L-1}}^N$, whose randomness stems from the random variables $\mathbf{h}^{L-1}$ and the random matrix $\mathbf{X}$.
Conditionally on  $K^N$, the corresponding Gaussian process $z_{K^N} \sim \mathcal{GP} (0, K^N)$ is well-defined and Bayesian inference can be performed with equations (\ref{nngp mean}-\ref{nngp variance}). 

\subsection{Limiting spectral distributions of NNGP kernel random matrices}

The following theorem establishes the relationship between the limiting spectral distribution of the \emph{actual} NNGP kernel random matrix $\mathbf{K}_{\mathbf{X},\mathbf{X}}$ and the \emph{empirical} NNGP kernel random matrix $\mathbf{K}_{\mathbf{X},\mathbf{X}}^{N}$. 

\begin{theorem} \label{deep non linear}
    Consider an NNGP obtained with the infinite-width limit of (\ref{random nn}) with $L \geq 2$, $N_L = 1$ and the non-linear activation $\phi$, a measurable, Lipschitz function. Consider the associated NNGP kernel denoted $K$, the associated random kernel function $K^{N}$ and the random matrix $\mathbf{K}_{\mathbf{X},\mathbf{X}}^{N}$ defined by Theorem \ref{random field of kernels} for kernel $K$. Then, the random matrix $\mathbf{K}_{\mathbf{X},\mathbf{X}}$ admits, in the limit $n, d \to \infty$ with fixed ratio $\frac{n}{d} = \psi \in (0, \infty)$, a limiting nonrandom spectral measure $\mu_\psi^\phi$.
    Furthermore, in the limit $N, n, d \to \infty$ with fixed ratio $\frac{n}{N} = \gamma \in (0, \infty)$, $\frac{n}{d} = \psi \in (0, \infty)$, the empirical spectral distribution of $\mathbf{K}_{\mathbf{X},\mathbf{X}}^{N}$ converges in distribution to the nonrandom measure $\rho_{MP}^{\gamma} \boxtimes \mu_\psi^\phi$.
\end{theorem}

The proof of Theorem \ref{deep non linear} relies on a recent result in random matrix theory (Theorem 1 in \cite{mervelede2}). The complete proof is provided in the appendix.
As a corollary, for deep linear networks, if the data covariance matrix $\mathbf{X} \mathbf{X}^T$ admits a limiting spectral distribution $\mu_\psi$, the limiting spectral distribution of $\mathbf{K}_{\mathbf{X},\mathbf{X}}^{N}$ is $\rho_{MP}^{\gamma} \boxtimes \mu_\psi$. In particular, if the covariance matrix is isotropic, then $\mu_\psi = \rho_{MP}^{\psi}$ and the limiting spectral distribution is the Marchenko-Pastur map of a Marchenko-Pastur distribution $\rho_{MP}^{\gamma} \boxtimes \rho_{MP}^{\psi}$.

Here, we have made an important distinction between the random matrices $\mathbf{K}_{\mathbf{X},\mathbf{X}}^{N}$ and $\mathbf{K}_{\mathbf{X},\mathbf{X}}$, which was not made in previous works \cite{spectrack}. Indeed, it is not the same thing to consider the NNGP kernel $K$, which appears \emph{after} the width of a neural network is taken to infinity, and its counterpart $K^{N}$ \emph{before} the width is taken to infinity (which should be called the \emph{empirical} NNGP kernel).

Theorem \ref{deep non linear} tells us how the spectral distribution of the empirical covariance matrix of the features created by the neural network (\ref{random nn}) depends on the \emph{actual} conjugate kernel of its associated NNGP.

The important fact to notice for the interpretation of the double-descent curve in neural networks is that, in the extremely overparametrised regime $\gamma \to 0$, the spectral distribution becomes that of the NNGP kernel itself. Indeed, the fixed-point equation (\ref{marchenkopasturfixedpoint}), which characterises the Marchenko-Pastur map of $\mu_\psi^\phi$, becomes:
\begin{equation}\label{mp converge dirac}
    \boxed{S (z) = \int \frac{1}{x - z} d\mu_\psi^\phi (x) = S_{\mu_\psi^\phi} (z).}
\end{equation}
In other words, spectrally, the neural network behaves like its corresponding NNGP in the extremely overparametrised regime. 
We will next see how the generalisation error of the corresponding GP and kernel regressions depend on this spectral distribution and reproduces the double-descent behavior.


\subsection{Double descent in the generalisation error of NNGPs and kernel regression}

We are now in position to study the generalisation error of the corresponding Gaussian process and kernel regressions. We will calculate the generalisation error of kernel regression with kernel $K^N$:
    \begin{equation}\label{gen error kernel N}
        E_{\mathcal{K}} (K^N) := \mathbb{E}_{\mathbf{x}, y, \mathbf{X},\mathbf{y}} \left( (\bar{\mu}^{K^N}_{\mathbf{X},\mathbf{y},\mathbf{x}} - y)^2 \right) 
    \end{equation}
and the generalisation error of Gaussian process regression with GP $z_{K^N}$:
\begin{equation}\label{gen error GP N}
   E_{\mathcal{GP}} (K^N) := \mathbb{E}_{\mathbf{x}, \mathbf{X},\mathbf{y}} \left( \mathbb{E}_{y, \bar{z}_{K^N}} ((\bar{z}_{K^N} - y)^2 | \mathbf{X},\mathbf{y},\mathbf{x}) \right) 
\end{equation}
where $\bar{\mu}^{K^N}_{\mathbf{X},\mathbf{y},\mathbf{x}}$ is the prediction mean of the Gaussian process regression with prior $z_{K^N}$, $\bar{z}_{K^N} | \mathbf{X}, \mathbf{y}, \mathbf{x}$ is the posterior distribution of Gaussian process regression with prior $z_{K^N}$, and the expectations are taken over the out-of-sample data and the training samples. Note that these predictions depends on the realisation of the random kernel function $K^N$. We study these generalisation errors when all quantities go to infinity and averaging over the random kernel using $n = \gamma N$, and $d = \frac{\gamma N}{\psi}$:
\begin{equation}\label{gen error limit}
   E_{\mathcal{K}} (\gamma, \psi) := \lim_{N \to \infty} \mathbb{E}_{K^{N}} \left(E_{\mathcal{K}} (K^{N}) \right) \qquad E_{\mathcal{GP}} (\gamma, \psi) := \lim_{N \to \infty} \mathbb{E}_{K^{N}} \left(E_{\mathcal{GP}} (K^{N}) \right)
\end{equation}
The following theorem highlights the dependence of the generalisation errors with some terms of interest that solely depend on the spectral measure that we studied in the previous section. The limits of these spectral measures will give us the double-descent behavior.
\begin{theorem}\label{gen error}
    Under the same assumptions as in our Theorem \ref{deep non linear}, the limiting generalisation errors $E_{\mathcal{K}} (\gamma, \psi)$ and $E_{\mathcal{GP}} (\gamma, \psi)$ can be expressed:
    \begin{equation}\label{kernel reg error eq}
        E_{\mathcal{K}} (\gamma, \psi) = D (\gamma, \psi) + C(\gamma, \psi)  g (\gamma, \psi)^2  + B (\gamma,\psi) g_2 (\gamma, \psi) + A (\gamma, \psi) g (\gamma, \psi) 
    \end{equation}
    \begin{equation}\label{gp reg error eq}
        \begin{split}
        E_{\mathcal{GP}} (\gamma, \psi) = \bar{D}(\gamma, \psi) + C(\gamma, \psi)  g (\gamma, \psi)^2 + B (\gamma, \psi) g_2 (\gamma, \psi) + \bar{A} (\gamma, \psi) g (\gamma, \psi) 
        \end{split}
    \end{equation}
where:
\begin{equation}\label{g function}
    g (\gamma, \psi) := \int_{0}^{\infty}  \frac{1}{\lambda+ \sigma_\epsilon^2} \mathrm{d} (\rho_{MP}^{\gamma} \boxtimes \mu_{\psi}^{\phi}) (\lambda) \qquad    g_2 (\gamma, \psi) := \int_{0}^{\infty}  \frac{1}{\left(\lambda + \sigma_\epsilon^2\right)^2} \mathrm{d} (\rho_{MP}^{\gamma} \boxtimes \mu_{\psi}^{\phi}) (\lambda)
\end{equation}
and $A(\gamma,\psi), \bar{A} (\gamma,\psi), B (\gamma,\psi), C (\gamma,\psi), D (\gamma,\psi), \bar{D} (\gamma,\psi)$ are bounded with respect to $n, N, d \to \infty$, and $B (\gamma,\psi)$ is non-zero.
\end{theorem}

The proof of Theorem \ref{gen error}, which is provided in the appendix, relies on the diagonalisation of the kernel random matrix $\mathbf{K}_{\mathbf{X},\mathbf{X}}^{N}$ and exploits Theorem \ref{deep non linear} to compute the expectation of the inverse of the eigenvalues in the limit of infinite quantities.  
These expressions capture the double descent behaviour as per the following corollary.

\begin{corollary}\label{dd nngp kernel}
Suppose that the assumptions of  Theorem \ref{deep non linear} hold true. Then, in the limit of $\sigma_\epsilon \to 0$ (noise-free), the generalisation error $E_{\mathcal{K}} (\gamma, \psi)$  exhibits a double descent with respect to $\gamma$. More precisely, the asymptote for the underparametrised regime is given by:
\begin{equation}\label{under regime}
    \lim_{\gamma \to \infty} E_{\mathcal{K}} (\gamma, \psi) = D (\gamma, \psi).
\end{equation} 
The asymptote for the interpolation threshold is given by:
\begin{equation}\label{interp thr 1}
    \lim_{\gamma \to 1^-} E_{\mathcal{K}} (\gamma, \psi) = \infty \qquad   \lim_{\gamma \to 1^+} E_{\mathcal{K}} (\gamma, \psi) = \infty.
\end{equation}
Finally the asymptote $\lim_{\gamma \to 0} E_{\mathcal{K}} (\gamma, \psi)$ for the overparametrised regime is finite and given by
\begin{equation*}\label{over regime}
    D(\gamma,\psi) +  C(\gamma, \psi) \left(\int_{0}^{\infty}  \frac{1}{\lambda} \mathrm{d} (\mu_{\psi}^{\phi}) (\lambda) \right)^2 + B(\gamma, \psi) \int_{0}^{\infty}  \frac{1}{\lambda^2} \mathrm{d} (\mu_{\psi}^{\phi}) (\lambda)  + A (\gamma,\psi) \int_{0}^{\infty}  \frac{1}{\lambda} \mathrm{d} (\mu_{\psi}^{\phi}) (\lambda)
\end{equation*}
The result also holds for $E_{\mathcal{GP}} (\gamma, \psi)$, replacing $A (\gamma,\psi), D (\gamma,\psi)$ by $\bar{A} (\gamma,\psi), \Bar{D} (\gamma,\psi)$.
\end{corollary}

We can see that the possibility of convergence to a finite value in the over-parametrised regime is enabled by the behavior of the Marchenko-Pastur map, as already explained by equation (\ref{mp converge dirac}). Indeed, the empirical spectral distribution converges to that of the \emph{actual} NNGP kernel matrix. 
The divergence at the interpolation threshold is due to eigenvalues becoming arbitrarily close to zero, due to a structural property independent of the input data distribution: the strictly positive support of the nonrandom measure $\rho_{MP}^{\gamma} \boxtimes \mu_{\psi}^{\phi}$ becomes arbitrarily close to zero when $\gamma \to 1$. In practice, the divergence is reduced by the effects of regularisation (in the case of NNGP regression, the noise model). More details are given in the appendix.

\section{Experiments}

In this section, we illustrate empirically that our results accurately predict the spectral distribution of NNGP kernel random feature matrices as well as the double-descent phenomenon of the generalisation errors of NNGP kernel regression. The experiments were conducted using GPU on Google Colab.

We have simulated the empirical spectral distribution of the kernel random matrix $\mathbf{K}_{\mathbf{X},\mathbf{X}}^N$ for high values of $N, n, d$ for ReLU and tanh with both a synthetic dataset, where the data is drawn from an isotropic multivariate Gaussian distribution $\mathbb{P}_d = \mathcal{N} (0, \frac{1}{d} I_d)$, and the MNIST dataset \cite{mnist}. 

\begin{figure}[ht]
\vskip 0.2in
\begin{minipage}{.5\textwidth}
\centerline{\includegraphics[height=4.5cm]{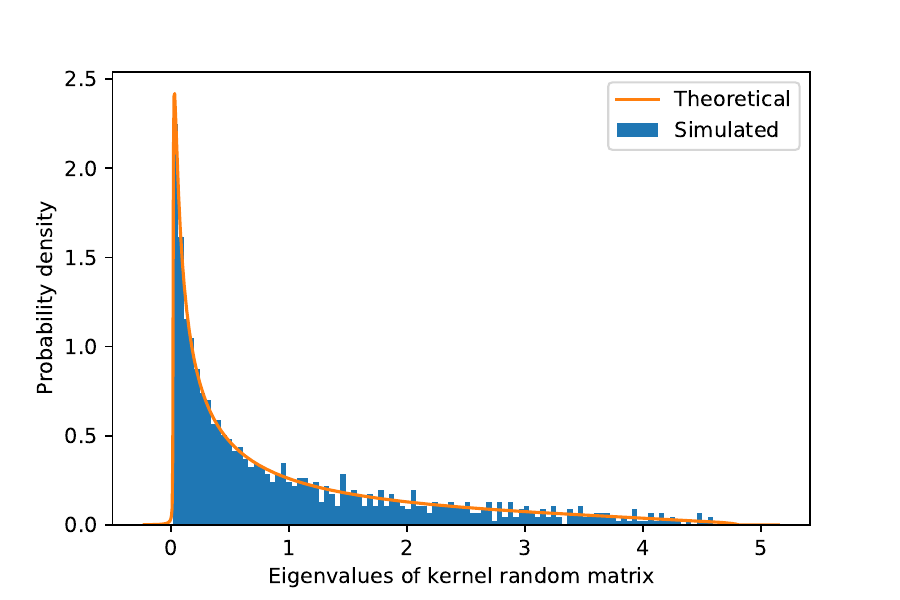}}
\end{minipage}
\begin{minipage}{.45\textwidth}
\centerline{\includegraphics[height=4.5cm]{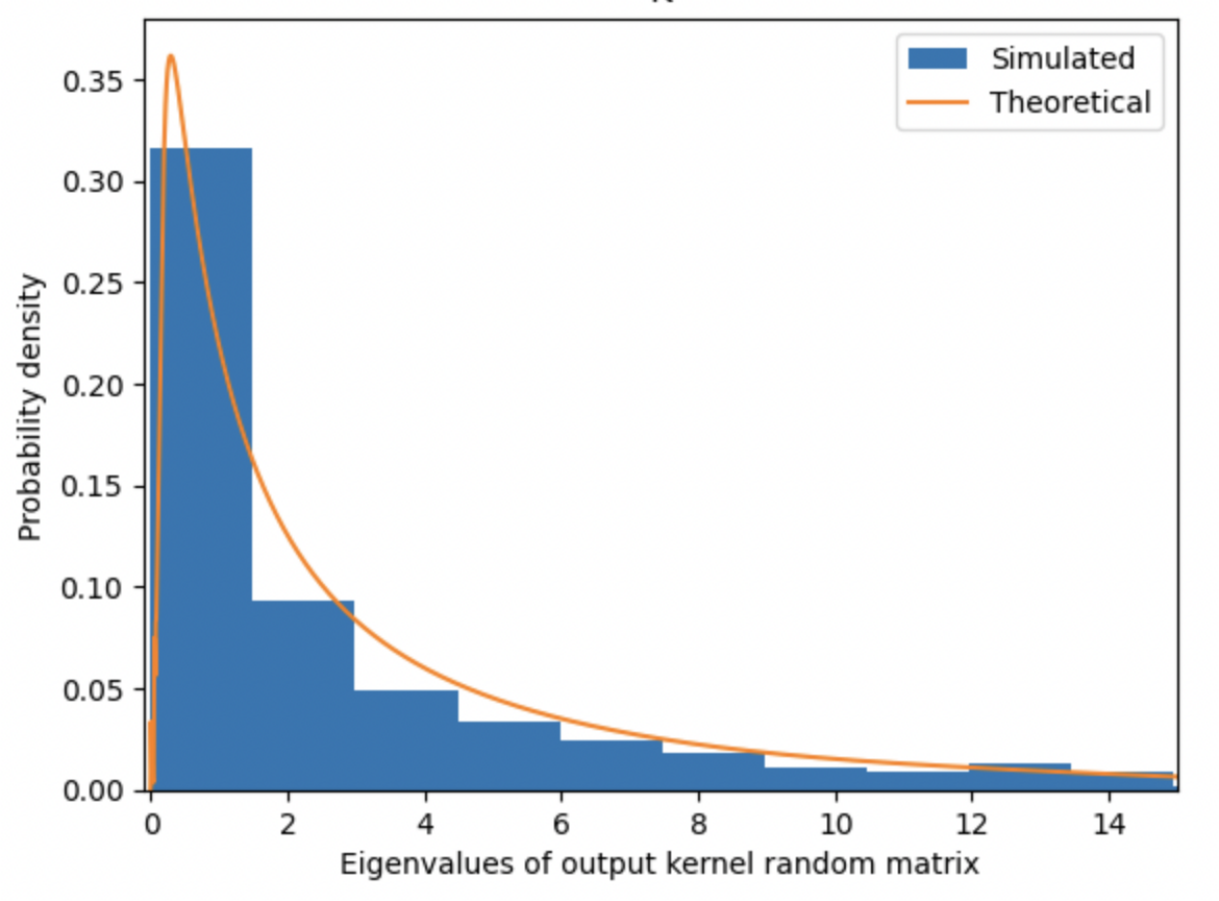}}
\end{minipage}
\caption{Simulated empirical spectral distribution versus theoretical limiting spectral distribution of the empirical covariance matrix $\mathbf{K}_{\mathbf{X},\mathbf{X}}^N$. On the left, we use a two-layer NNGP with no non-linearity 
 with teacher distribution $\mathcal{N} (0, \frac{1}{d} I_d)$ using $N = 300$, $n = 200$, and $d = 400$. On the right, we use a two-layer ReLU NNGP on a subset of MNIST taking  $N = 600$, $n = 300$, and $d = 784$ (number of pixels on MNIST images).The simulated distribution is obtained by sampling from the random matrix, and the theoretical distibution is obtained by solving the Marchenko-Pastur fixed-point equation.}
\label{eigenvaluedistribsynth}
\end{figure}

As illustrated in Figure \ref{eigenvaluedistribsynth}, we found an excellent good agreement with the theoretical prediction of the limiting spectral distributions. We used the Marchenko-Pastur fixed point equation \eqref{marchenkopasturfixedpoint} to compute the limiting spectral distribution $\rho_{MP}^{\gamma} \boxtimes \mu_\psi^\phi$, by iterating over the recursive sequence it defines in the Stieltjes transform space and then inverting the Stieltjes transform using the inversion formula. In the case of synthetic data drawn from $\mathbb{P}_d = \mathcal{N} (0, \frac{1}{d} I_d)$ and with no nonlinearity, the actual NNGP kernel matrix can be characterised exactly by $\mu_\psi^\phi = \rho_{MP}^{\psi}$. In the case of MNIST with ReLU, the actual NNGP kernel is not known, hence we estimated the actual NNGP kernel matrix by sampling $\mathbf{K}_{\mathbf{X},\mathbf{X}}^{\hat{N}}$ with a very large value of $\hat{N} \gg n$, i.e.\ precisely relying on the fact that $\rho_{MP}^{0} \boxtimes \mu_\psi^\phi = \mu_\psi^\phi$. We focused on a subset of MNIST restricted to digits "0" and "1" in order to simplify the structure of the covariance matrices and their spectral distributions; this procedure provides an excellent agreement.

\begin{figure}[ht]
\vskip 0.2in
\begin{minipage}{.48\textwidth}
\centerline{\includegraphics[height=4.5cm]{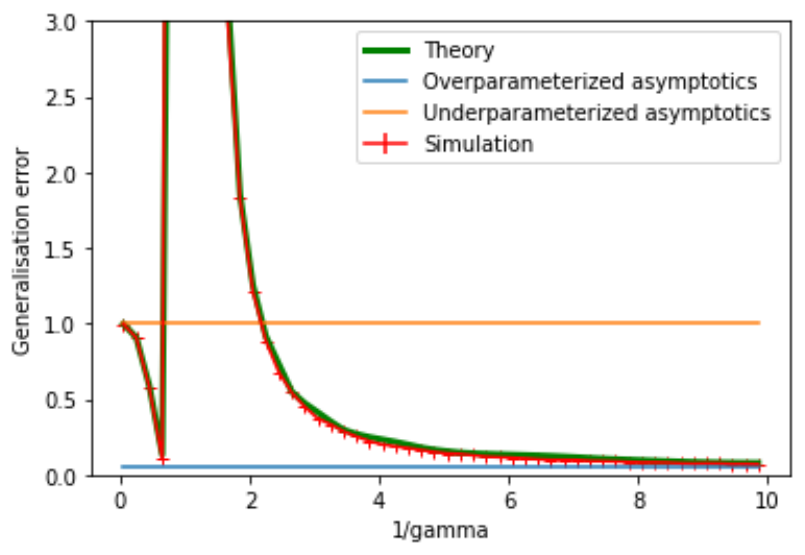}}
\end{minipage}
\begin{minipage}{.48\textwidth}
\centerline{\includegraphics[height=4.5cm]{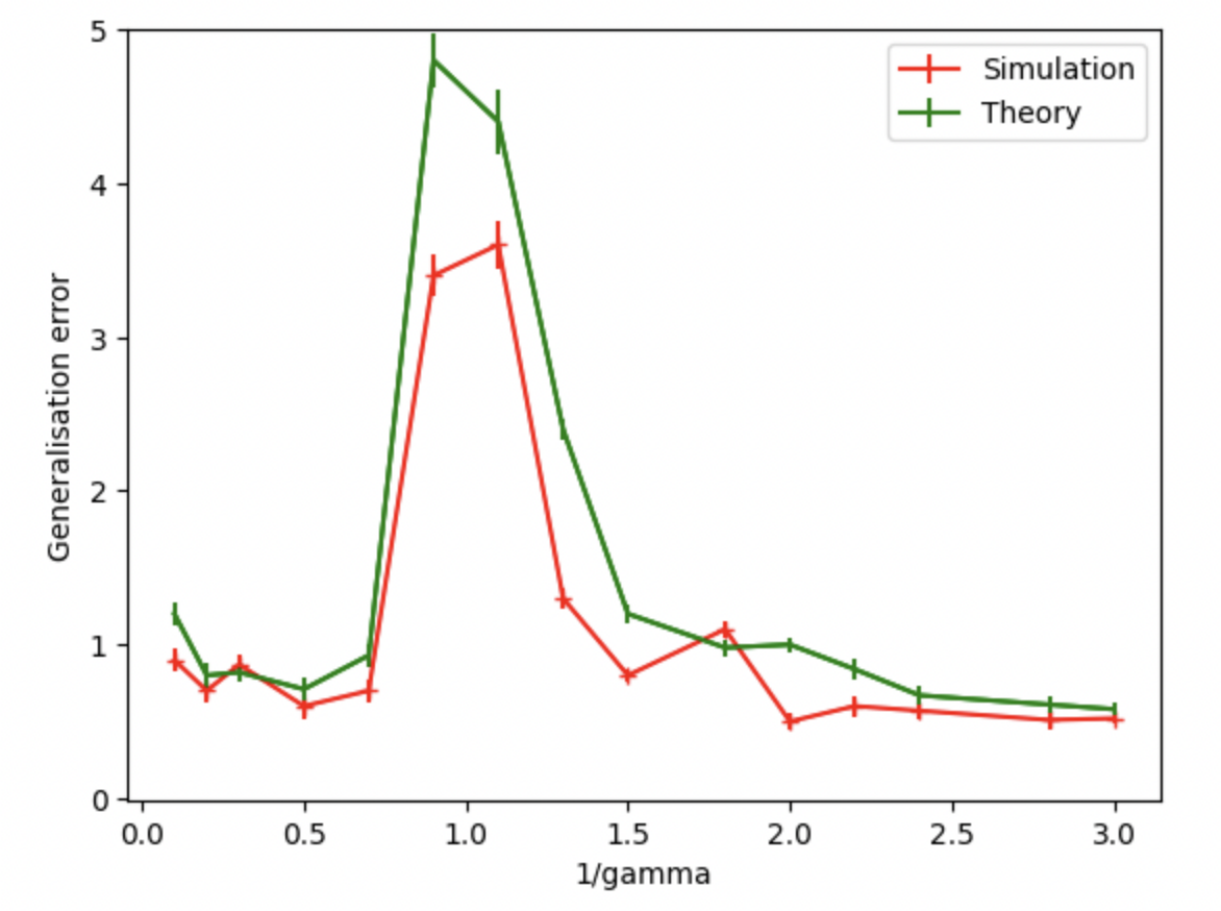}}
\end{minipage}
\label{generrorfigsynth}
\caption{Simulated generalisation error versus theoretical limiting generalisation error, as a function of $\frac{1}{\gamma}$.  On the left, the simulated error is obtained by sampling from the kernel random matrix of a two-layer NNGP with no non-linearity under teacher distribution $\mathcal{N} (0, \frac{1}{d} I_d)$, and the theoretical distibution and the theoretical asymptotes are obtained by integrating $\frac{1}{\lambda}$ and $\frac{1}{\lambda^2}$ over the Marchenko-Pastur map of the distribution $\rho_{MP}^{\psi}$. On the right, the simulated error is obtained by sampling from the kernel random matrix of a two-layer ReLU NNGP on a subset of MNIST, and the theoretical distribution is obtained by sampling from the Marchenko-Pastur map of the empirical NNGP kernel matrix $\mathbf{K}_{\mathbf{X},\mathbf{X}}^{\hat{N}}$, and from independent Gaussian distributions in lieu of eigenfunctions.}
\end{figure}

We have simulated the generalisation errors of NNGP kernel regression on the same datasets. To calculate the generalisation errors, we relied on the spectral universality assumption (SUA) \cite{Sollich_2002} to estimate eigenfunctions (and hence coefficients $A(\gamma,\psi), B (\gamma,\psi), C (\gamma,\psi), D (\gamma,\psi)$), which states that in high dimension eigenfunctions become unstructured and can be approximated by independent Gaussian entries. Determining in which cases the SUA is valid is still an active area of research \cite{El_Karoui_2010,doi:10.1142/S201032631350010X,Fan2015TheSN,kernelreghighdim,lu2023equivalence}. In the case of isotropic data and no nonlinearity, the SUA is exact \cite{El_Karoui_2010}, which allows us to find a very good agreement (Figure \ref{generrorfigsynth}). In the case of MNIST with ReLU, there is evidence that the SUA does apply to some extent, as demonstrated by \cite{simon2022eigenlearning} who also uses SUA to estimate generalisation errors of kernel regression in high dimensions on MNIST.  We found an acceptable agreement between theory and simulations. The main sources of discrepancies 
stem from the fact that we only use a small subset (300 examples) to estimate the empirical spectral distribution and that the SUA may not be completely accurate in this particular setting. Although our predictions for the generalisation error are not perfectly accurate, we emphasise that they correctly predict the double-descent phenomenon, and thus support our claim that the double-descent phenomenon is only driven by the spectral distribution and its dependence with the width.

\section{Related Work}

The properties of stochastic gradient descent (SGD) have been proposed as an explanation for the favourable generalisation power of DNNs in the over-parametrised regime; for instance, the tendency to escape saddle points \cite{sgdsaddles}  could explain how  solutions that generalise well are selected over all others. The neural tangent kernel describes the dynamics of SGD in the functional space and its relationship to the generalisation power of
DNNs is well documented \cite{ntk,ntktriple,specdecntk,spectralbias,statsmech}.

The favourable generalisation properties of DNNs in the over-parametrised regime may also be explained using Bayesian methods and other kernel machines in a way that is unrelated to SGD training; indeed, the good performance of NNGP regression \cite{nngp,lee2020finite} provides compelling evidence in this direction. A related argument is that the parameter-function map is exponentially biased towards Kolmogorov simple functions~\cite{dingle2018input,valle2018deep,mingard2019neural}; since the data on which DNNs are trained has structure, this inductive bias leads to good generalisation in the over-parametrised regime. Due to large differences in the sizes of the basins of attraction~\cite{schaper2014arrival}, SGD converges to functions with a probability that is remarkably close to the Bayesian posterior probability that a DNN expresses upon random sampling of parameters~\cite{mingard2020sgd}.  These ideas are still being actively debated \cite{nnvskernel, bayesiandeep,belkin2021fit}.

The seminal work of \cite{Vallet1989dd, sompolinsky} on the double-descent phenomenon and the subsequent developments in  \cite{randomfeature, surprises, highdim, kernelreghighdim} suggest that the
favourable generalisation power of DNNs is an intrinsic characteristic of the set of functions that these models can learn, as generalisation errors are computed analytically and independently from any learning algorithm. The methods of Statistical Physics have traditionally been the tool of choice for obtaining closed-form formulae in this setting \cite{statmechoflearning}. In this line of research, recent works have provided analytical expressions for the generalisation error of high-dimensional kernel regressions \cite{specbiasandtaskalign,specdecntk,NEURIPS2020_b367e525,simon2022eigenlearning,Cui_2022}. In particular, \cite{NEURIPS2020_b367e525} and \cite{simon2022eigenlearning} rely on the spectral universality assumption, just as we do to estimate the coefficients in our formula. As pointed by \cite{simon2022eigenlearning}, other works take the spectral universality assumption implicitly via, for instance, the replica method \cite{specdecntk,specbiasandtaskalign}. Our computation of the generalization error is thus similar to the works of \cite{NEURIPS2020_b367e525,simon2022eigenlearning}. Their results however hold for frozen kernels and the dependence with the width is not studied.

 
The limiting spectral distributions of the kernel random matrices that we study in this paper were first investigated in \cite{spectrack}. Our results are, however, stronger since they require less restrictive assumptions on the data generating process and the non-linear activations; for instance, we do not assume the non-linear activation to be twice differentiable nor the columns of the input data matrix to be "$(\epsilon, B)$-orthonormal". This was made possible by deriving the analytical expression as a function of an implicit quantity: the spectral measure of the actual NNGP kernel. Furthermore, we emphasise that the link between the "CK" kernel random matrix in \cite{spectrack}  and the actual conjugate (NNGP) kernel is not straightforward. This subtle distinction enables a more transparent interpretation of the the double-descent phenomenon.

The double-descent behaviour in the learning curves of high-dimensional kernel regression (including the NNGP and neural tangent kernels as particular cases) has been described in
\cite{specbiasandtaskalign, specdecntk}. Our work improves on this line of research by introducing the idea of width-dependent kernels, which is especially well-suited to the context of DNNs where double descent manifests as the network width tends to infinity.
Recent studies of the double-descent phenomenon have focused on random features regressions in the case of shallow networks \cite{asymptoticserror,randomfourierfeat,generalizedlinear,generalizedhidden,benign,finegrainedd,twomodelsweakfeatures,conditionrandom,convexregularization}, or kernel regression with no dependence on the width \cite{kernelreghighdim,taxonomyoverfitting}. 


\section{Conclusions}

In this paper, we have exploited results from random matrix theory to offer a new perspective on the double descent phenomenon in FCNs through the lens of Gaussian process kernels. We have derived analytical expressions for the generalisation error under teacher-student scenarios, which are applicable to networks of arbitrary depth and a large family of nonlinearities. This analysis allows us to predict the double descent behaviour as the width of the last layer changes relatively to the number of examples, and understand it as simply arising from the discrepancy between the spectrum of width-dependent random kernel (corresponding the empirical covariance matrix of the features), and that of the width-independent NNGP kernel. Finally, we hope that the tools we have developed will motivate further research on the properties of the generalisation error of neural networks.

\paragraph{Limitations.}  Although our approach provides a transparent interpretation for the double-descent phenomenon, it may not always be accurate in quantifying the generalisation error. Indeed, our computations relies on the spectral universality assumption, which may not be fully valid in real-world datasets. Another limitation of our approach is that the theory currently applies only to fully-connected neural architectures and assumes infinite dimensions.

\printbibliography
\clearpage

\section{Proof of Proposition 1.}

The proof is immediate using that the fact the $\mathbf{h}_i^{L-1}$ are independent, $\mathbf{h}_i^{L-1} \sim \mathcal{GP} (0, K^{\phi, L-1})$, and that $\frac{1}{N} \sum_{i=1}^N A_i$ is an unbiased estimator of $\mathbb{E} (A)$ with variance $\frac{\mathrm{Var} (A)}{N}$ for any collection of random variables $A_i, A$ such that $A_i \sim A$ sampled i.i.d.\ . To calculate the variance, in order to show that it is $\mathcal{O}_{N \to \infty} (1)$ (in standard big O notation), one can use the same trick as in \cite{poole}, where they computed an iterative map of the covariances in the limit of infinite width:
\begin{equation} \label{iterative map}
        K^l (\mathbf{x}, \mathbf{x}') = \int_{-\infty}^\infty \int_{-\infty}^\infty \phi\left( u_1^{\phi, l-1} (\mathbf{x},\mathbf{x}')\right) \phi\left(u_2^{\phi, l-1} (\mathbf{x},\mathbf{x}') \right) \mathcal{D}z_1 \mathcal{D}z_2
\end{equation}
and 
\begin{equation} \label{intial cov}
    K^{\phi,0} (\mathbf{x}, \mathbf{x}') = \langle \mathbf{x}, \mathbf{x}' \rangle
\end{equation}
where $\mathcal{D}z$ is the standard Gaussian measure, and 
\begin{equation}
    u_1^{\phi, l-1} (\mathbf{x},\mathbf{x}') = \sqrt{K^{\phi, l-1} (\mathbf{x},\mathbf{x})} z_1
\end{equation}
\begin{equation}
    u_2^{\phi, l-1} (\mathbf{x},\mathbf{x}') = \sqrt{K^{\phi, l-1} (\mathbf{x}',\mathbf{x}')} \left[ \rho^{\phi, l-1} (\mathbf{x},\mathbf{x}') z_1 + \sqrt{1 - \rho^{\phi, l-1} (\mathbf{x},\mathbf{x}')^2} z_2 \right]
\end{equation}
with $\rho^{\phi, l-1} (\mathbf{x},\mathbf{x}') = \frac{K^{\phi, l-1}(\mathbf{x},\mathbf{x}')}{\sqrt{K^{\phi, l-1}(\mathbf{x},\mathbf{x})K^{\phi, l-1}(\mathbf{x}',\mathbf{x}')}}$. The trick is that $(z_1, z_2)$ is the joint density of independent standard Gaussians, and
$\begin{pmatrix}
    u_1^{\phi, l-1} (\mathbf{x},\mathbf{x}') \\ u_2^{\phi, l-1} (\mathbf{x},\mathbf{x}')
\end{pmatrix}$
is thus the density of a bivariate Gaussian with covariance matrix 
$\begin{pmatrix} 
    K^{\phi, l-1} (\mathbf{x},\mathbf{x}) & K^{\phi, l-1} (\mathbf{x},\mathbf{x}') \\ K^{\phi, l-1} (\mathbf{x}',\mathbf{x}) & K^{\phi, l-1} (\mathbf{x}',\mathbf{x}')
\end{pmatrix}$.  We find that:
\begin{equation}
    \mathrm{Var} \left(\left( \mathbf{K}_{\mathbf{X}, \phi, \mathbf{h}^{L-1}}^N \right)_{i,j} \right) = \frac{1}{N} \left[\int_{-\infty}^\infty \int_{-\infty}^\infty \phi\left( u_1^{\phi, L-1} (\mathbf{x}_i,\mathbf{x}_j)\right)^2 \phi\left(u_2^{\phi, L-1} (\mathbf{x}_i,\mathbf{x}_j) \right)^2 \mathcal{D}z_1 \mathcal{D}z_2 - K^{\phi, L} (\mathbf{x}_i, \mathbf{x}_j)^2\right]
\end{equation}

\section{Proof of Theorem 1.}
The trick is to utilise a stochastic process which provides a probability space where one can sample from infinitely-, uncountably-, many Gaussian distributions. Here we chose the Brownian motion.

Consider the probability space $(\Omega_N, \mathcal{A}_N, \mathbb{P}_N)$ associated with an $N$-dimensional standard Brownian motion, $(B_t^k)_{t\in \mathbb{R}}$ for $k \in [N]$. Define the following random variable in $\mathbb{R}^{(\mathbb{R}^d)^2}$:
\begin{equation}
    \begin{split}
    K^{\phi, L, N} : \;& \Omega_N \longrightarrow \mathbb{R}^{(\mathbb{R}^d)^2}   \\ & \omega \mapsto K^{\phi, L, N} (\omega) 
    \end{split}
\end{equation}
with
\begin{equation}
    \begin{split}
        K^{\phi, L, N} (\omega) : \;& (\mathbb{R}^d)^2 \rightarrow \mathbb{R}   \\ & (\mathbf{x}, \mathbf{x}') \mapsto  \frac{1}{N} \sum_{k = 1}^N \phi \left( u_1^{\phi, L-1, B^k} (\mathbf{x},\mathbf{x}') (\omega) \right) \phi \left( u_2^{\phi, L-1, B^k} (\mathbf{x},\mathbf{x}') (\omega) \right)
    \end{split}
\end{equation}
where
\begin{equation}
    u_1^{\phi, L-1, B^k} (\mathbf{x},\mathbf{x}') = B_{K^{\phi, L-1} (\mathbf{x},\mathbf{x})}^k
\end{equation}
and 
\begin{equation}
\begin{split}
    u_2^{\phi, L-1, B^k} (\mathbf{x},\mathbf{x}') = & \frac{\sqrt{K^{\phi, L-1} (\mathbf{x},\mathbf{x}')}}{K^{\phi, L-1} (\mathbf{x},\mathbf{x})} B_{K^{\phi, L-1} (\mathbf{x},\mathbf{x})}^k \\& + \sqrt{1 - \frac{K^{\phi, L-1} (\mathbf{x},\mathbf{x}')} {K^{\phi, L-1} (\mathbf{x},\mathbf{x}) K^{\phi, L-1} (\mathbf{x}',\mathbf{x}')}} \left(B_{K^{\phi, L-1} (\mathbf{x},\mathbf{x})+ K^{\phi, L-1} (\mathbf{x}',\mathbf{x}')}^k - B_{K^{\phi, L-1} (\mathbf{x},\mathbf{x})}^k \right)
\end{split}
\end{equation}

Using the same technique introduced by \cite{poole} for the iterative map of the covariances in the conjugate kernel (see equation (\ref{iterative map})), we have that $\left(u_1^{\phi, L-1,B^k} (\mathbf{x},\mathbf{x}'), u_2^{\phi, L-1,B^k} (\mathbf{x},\mathbf{x}')\right)$ follows a bivariate Gaussian with covariance matrix $\begin{pmatrix}
    K^{\phi, L-1} (\mathbf{x},\mathbf{x}) & K^{\phi, L-1} (\mathbf{x},\mathbf{x}') \\ K^{\phi, L-1} (\mathbf{x}',\mathbf{x}) & K^{\phi, L-1} (\mathbf{x}',\mathbf{x}')
\end{pmatrix}$, where we have used that $B_{K^{\phi, L-1} (\mathbf{x},\mathbf{x})}^k \sim \mathcal{N} (0, K^{\phi, L-1} (\mathbf{x},\mathbf{x}))$, and  $B_{K^{\phi, L-1} (\mathbf{x},\mathbf{x})+ K^{\phi, L-1} (\mathbf{x}',\mathbf{x}')}^k - B_{K^{\phi, L-1} (\mathbf{x},\mathbf{x})}^k \sim \mathcal{N} (0, K^{\phi, L-1} (\mathbf{x}',\mathbf{x}'))$ are independent, by properties of the Brownian motion. We have that $\left( \mathbf{h}_k^{L-1} (\mathbf{x}), \mathbf{h}_k^{L-1} (\mathbf{x}') \right)$ follows the same bivariate Gaussian distribution since $\mathbf{h}_k^{L-1} \sim \mathcal{GP} (0, K^{\phi, L-1})$, thus the real-valued random variable $\omega \mapsto K^{\phi, L, N} ( \omega) (\mathbf{x}_i, \mathbf{x}_j)$ has the same distribution as $\left(\mathbf{K}_{\mathbf{X}, \phi, \mathbf{h}^{L-1}}^N \right)_{i,j}$, for all $i, j$ which gives us the point 2. 

Equation \eqref{kernel property} thus proves that $K^{\phi, L, N} ( \omega)$, which is a well-defined function of $\mathbb{R}^{(\mathbb{R}^d)^2}$, is a kernel, which terminates the proof.

\section{Proof of Theorem 2.}

For clarity, we first prove the result for a two-layer linear network with isotropic input data, then we extend to a deep linear network with our mild assumption on the data distribution, and finally we generalize the result to a deep non-linear network under our mild assumptions on the nonlinearity. 

\subsection{Two-layer linear network with isotropic input data.}

We consider a two-layer linear network $L=2$, $\phi = \mathrm{Id}_{\mathbb{R}}$, and isotropic input data $\mathbb{P}_d = \mathcal{N} (0, \frac{1}{d} \mathbf{I}_d)$.

Let $N, n, d \in (0, \infty)$ such that $\frac{n}{d^\alpha} = \psi \in (0, \infty)$, $\frac{N}{n} = \gamma \in (0, \infty)$. 

Conditionally on $\mathbf{X}$, $\mathbf{h}_1^1, ..., \mathbf{h}_N^1$ are $N$ independent realisations of the $n$-dimensional multivariate Gaussian $\mathcal{N} (0, \mathbf{X}\mathbf{X}^T)$ as implied by equation (\ref{intial cov}). We will note $\mathbf{H}_\mathbf{X} = (\mathbf{h}_1^{1}, ..., \mathbf{h}_N^{1})^T \in \mathbb{R}^{N \times n}$, and we have $\mathbf{K}_{\mathbf{X}, \phi, \mathbf{h}^{1}}^N = \frac{1}{N} \mathbf{H}_\mathbf{X}^T \mathbf{H}_\mathbf{X}$ as per Proposition \ref{unbiased estimator}.

The idea is to diagonalise the symmetric matrix $\mathbf{X}\mathbf{X}^T$ in order to deal with uncorrelated random variables: 
\begin{equation}
    \mathbf{X}\mathbf{X}^T = \mathbf{V}_\mathbf{X} \mathbf{\Lambda}_\mathbf{X} \mathbf{V}_\mathbf{X}^T
\end{equation}
where $\mathbf{\Lambda}_\mathbf{X}$ is a diagonal matrix, $\mathbf{V}_\mathbf{X}$ is an orthonormal matrix.

Now, consider $\hat{\mathbf{h}}_1, ..., \hat{\mathbf{h}}_N$, $N$ independent realisations of $\mathcal{N}(0, \mathbf{\Lambda}_\mathbf{X})$, and define $\hat{\mathbf{H}}_\mathbf{X} = (\hat{\mathbf{h}}_1, ..., \hat{\mathbf{h}}_N)^T \in \mathbb{R}^{N \times n}$. We have that $\mathbf{H}_\mathbf{X}$ and $\hat{\mathbf{H}}_\mathbf{X} \mathbf{V}_\mathbf{X}^T$ have the same distribution conditionally on $\mathbf{X}$, thanks to the linear properties of the Gaussian. This implies that $\mathbf{H}_\mathbf{X}^T \mathbf{H}_\mathbf{X}$ and $\mathbf{V}_\mathbf{X} \hat{\mathbf{H}}_\mathbf{X}^T \hat{\mathbf{H}}_\mathbf{X} \mathbf{V}_\mathbf{X}^T$ have the same distribution conditionally on $\mathbf{X}$, thus the same distribution when the conditioning is relaxed. In particular, they have the same spectral distribution. We have reduced the problem to studying the existence of a limiting spectral distribution for $\frac{1}{N} \mathbf{V}_\mathbf{X} \hat{\mathbf{H}}_\mathbf{X}^T \hat{\mathbf{H}}_\mathbf{X} \mathbf{V}_\mathbf{X}^T$.

Furthermore, $\frac{1}{N}\hat{\mathbf{H}}_\mathbf{X}^T \hat{\mathbf{H}}_\mathbf{X}$ and $\frac{1}{N} \mathbf{V}_\mathbf{X} \hat{\mathbf{H}}_\mathbf{X}^T \hat{\mathbf{H}}_\mathbf{X} \mathbf{V}_\mathbf{X}^T$ have the same spectral distribution, since they only differ by a basis change. We have thus further reduced the problem to studying the existence of a limiting spectral distribution for $\frac{1}{N}\hat{\mathbf{H}}_\mathbf{X}^T \hat{\mathbf{H}}_\mathbf{X}$.

As per the definition of $\hat{\mathbf{H}}_\mathbf{X}$, it can be re-written:
\begin{equation}\label{rewrittingh}
    \hat{\mathbf{H}}_\mathbf{X} =  \sqrt{N} \mathbf{M}\mathbf{\Lambda}_\mathbf{X}^{1/2}
\end{equation}
where  $\mathbf{M}= (\mathbf{m}_1, ..., \mathbf{m}_N)^T \in \mathbb{R}^{N \times n}$ with $\mathbf{m}_1, ..., \mathbf{m}_N$ $N$ independent realisations of $\mathcal{N}(0, \frac{1}{N} \mathbf{I}_n)$.

In the limit of infinite quantities, since the data distribution is isotropic, the limit of the spectral measure of $\mathbf{X}\mathbf{X}^T$ exists almost surely: it is the Marchenko-Patur distribution $\rho_{MP}^{\psi}$ \cite{marchenkopastur}.

Conditionally on $\mathbf{X}$, $\mathbf{\Lambda}_\mathbf{X}$ is deterministic and positive semi-definite (diagonal positive), thus the result on the Marchenko-Pastur map applies (equation \eqref{marchenkopasturfixedpoint}), and $\frac{1}{N}\hat{\mathbf{H}}_\mathbf{X}^T \hat{\mathbf{H}}_\mathbf{X} = \left(\mathbf{\Lambda}_\mathbf{X}^{1/2}\right)^T \mathbf{M}^T\mathbf{M}\mathbf{\Lambda}_\mathbf{X}^{1/2}$ admits a limiting spectral distribution: $\rho_{MP}^{\gamma} \boxtimes \mu_\mathbf{X}$, where  $\mu_\mathbf{X}$ is the infinite limit of the spectral measure of a given realisation of $\mathbf{X}\mathbf{X}^T$ (because of the conditioning), which exists almost surely.

Relaxing the conditioning on $\mathbf{X}$ gives us the limiting spectral distribution $\rho_{MP}^{\gamma} \boxtimes \rho_{MP}^{\psi}$, because $\mu_\mathbf{X}$ is almost surely $\rho_{MP}^{\psi}$. In other words, the empirical spectral distribution converges almost surely to $\rho_{MP}^{\gamma} \boxtimes \rho_{MP}^{\psi}$, which terminates the proof for a two-layer linear network with isotropic input data.

\subsection{Deep linear network and input data with bounded dot product.}

We now consider a deep linear network $\phi = \mathrm{Id}_{\mathbb{R}}$ and a data distribution $\mathbf{x}\sim \mathbb{P}_d$ such that $\mathbb{E}_{\mathbf{x} \sim \mathbb{P}_d} (\mathbf{x}^T \mathbf{x})$ is bounded as $d \to \infty$.

The extension of the proof is immediate following closely the previous proof, since we only used that the limiting spectral distribution of the empirical covariance matrix $\mathbf{X}\mathbf{X}^T$ is a Marchenko-Pastur distribution in the last stage, when relaxing the conditioning on $\mathbf{X}$. The assumption on the data distribution ensures that $\mathbf{X} \mathbf{X}^T$ admits a nonrandom limiting spectral measure $\mu_\psi$ in the limit $n,d \to \infty$, with $\frac{n}{d} = \psi$ \cite{Livan_2018}.

Furthermore, it can be shown by immediate induction, that when $\phi = \mathrm{Id}_\mathbb{R}$, the iterative map of the covariances of the conjugate kernel (see equation (\ref{iterative map})) yields $K^l (\mathbf{x}, \mathbf{x}') = \langle \mathbf{x}, \mathbf{x}' \rangle$ for all $l \geq 2$, thus $K_{\mathbf{X},\mathbf{X}}^{\phi, l, N} = \mathbf{X} \mathbf{X}^T$ for all $l \geq 2$. Since $\mathbf{h}_1^{L-1}, ..., \mathbf{h}_N^{L-1}$ are $N$ independent realisations of $\mathcal{N} (0, K_{\mathbf{X},\mathbf{X}}^{\phi, L-1,N})$, we can thus reproduce the previous proof with $\mathbf{H}_\mathbf{X}^{L-1} = (\mathbf{h}_1^{L-1}, ..., \mathbf{h}_N^{L-1})^T \in \mathrm{R}^{N \times n}$, and obtain the same limiting spectral distribution for $K_{\mathbf{X},\mathbf{X}}^{\phi, L, N}$.

\subsection{Deep non-linear network with measurable Lipschitz nonlinearity.}

First, let us briefly discuss why $K^{\phi, L}_{\mathbf{X},\mathbf{X}}$ has indeed a limiting nonrandom spectral measure $\mu^\psi_\phi$. 
Using that $\phi$ is $c$-Lipschitz, we have that:
\begin{equation}
    \begin{split}
    |\mathbb{E}_{\mathbf{x}\sim \mathbb{P}_d, f \sim \mathcal{GP} (0, K^0)} (\phi (f(\mathbf{x}))^2)| & \leq c^2 \; \mathbb{E}_{\mathbf{x}\sim \mathbb{P}_d, f \sim \mathcal{GP} (0, K^0)} ({|f(\mathbf{x})|}^2) \\& = c^2 \; \mathbb{E}_{\mathbf{x}\sim \mathbb{P}_d} (\mathbf{x}^T \mathbf{x}) \\& < \infty
    \end{split}
\end{equation}
By immediate induction on the depth $L$, since $\mathbb{E}_{\mathbf{x}\sim \mathbb{P}_d, f \sim \mathcal{GP} (0, K^0)} (\phi (f(\mathbf{x}))^2) < \infty$, it follows that $\mathbb{E}_{\mathbf{x},\mathbf{x}'} \left(K^{\phi,L} (\mathbf{x},\mathbf{x}')\right) < \infty$ (by applying successively Cauchy-Schwarz inequalities). This is sufficient to apply Mercer's decomposition theorem \cite{merceroriginal,mercertheorem}, which provides eigenvalues independent from the sampling effects of the data, and therefore a nonrandom spectral measure. One can conclude by invoking \cite{baker,gpforml} which proves that the eigenvalues of the empirical kernel matrix $K^{\phi, L}_{\mathbf{X},\mathbf{X}}$ converge to the Mercer's eigenvalues.\\

With a non-linear activation, the trick of diagonalising the covariance matrix $\mathbf{X}\mathbf{X}^T$ does not directly apply, since we need to consider $\phi (\mathbf{H}_\mathbf{X}^{L-1})$, where $\phi$ acts entrywise, and we cannot ensure that  $\phi (\mathbf{H}_\mathbf{X}^{L-1})$ has the same distribution as $\phi (\hat{\mathbf{H}}_\mathbf{X}^{L-1}) \mathbf{V}_\mathbf{X}^{L-1}$. Instead, we diagonalise the conjugate kernel matrix $K_{\mathbf{X},\mathbf{X}}^{\phi,L}$ in order to utilise a somewhat similar property. 

With the previous notation, we have that $K_{\mathbf{X},\mathbf{X}}^{\phi,L,N} = \frac{1}{N} \phi (\mathbf{H}_\mathbf{X}^{L-1})^T \phi (\mathbf{H}_\mathbf{X}^{L-1})$. We diagonalise the symmetric matrix $K_{\mathbf{X},\mathbf{X}}^{\phi,L}$.
\begin{equation}
    K_{\mathbf{X},\mathbf{X}}^{\phi,L} =  \mathbf{U}_\mathbf{X}^{\phi,L} \mathbf{\Lambda}_\mathbf{X}^{\phi,L} ( \mathbf{U}_\mathbf{X}^{\phi,L})^T 
\end{equation}
where $\mathbf{\Lambda}_\mathbf{X}^{\phi,L}$ is a diagonal matrix, and $ \mathbf{U}_\mathbf{X}^{\phi,L} \in \mathbb{R}^{n \times n}$ is an orthonormal matrix.
Consider, as previously, $\hat{\mathbf{h}}_1^{\phi,L}, ..., \hat{\mathbf{h}}_N^{\phi,L}$, N independent realisations of $\mathcal{N} (0, \mathbf{\Lambda}_\mathbf{X}^{\phi,L})$ and note $\hat{\mathbf{H}}_\mathbf{X}^{\phi,L} = (\hat{\mathbf{h}}_1^{\phi,L}, ..., \hat{\mathbf{h}}_N^{\phi,L})^T \in \mathbb{R}^{N \times n}$. In this case, $\hat{\mathbf{H}}_\mathbf{X}^{\phi,L} ( \mathbf{U}_\mathbf{X}^{\phi,L})^T$ and $\phi( \mathbf{H}_\mathbf{X}^{L-1})$ do not have the same distribution, because the distributions are not Gaussian anymore, but they have the same covariance structure, conditionally on $\mathbf{X}$:
\begin{equation}\label{covariancestructure}
    \begin{split}
    \mathbb{E}_{h} \left( \phi ( (\mathbf{h}_i^{L-1})_k ) \phi ( (\mathbf{h}_i^{L-1})_l ) | \mathbf{X}\right) & = \left( K_{\mathbf{X},\mathbf{X}}^{\phi,L} \right)_{k,l} \\ & = \sum_{p=1}^n \left( \mathbf{U}_\mathbf{X}^{\phi,L} \right)_{k,p} \left(\mathbf{\Lambda}_\mathbf{X}^{\phi,L} ( \mathbf{U}_\mathbf{X}^{\phi,L})^T) \right)_{p,l} \\ & = \sum_{p=1}^n \left( \mathbf{U}_\mathbf{X}^{\phi,L} \right)_{k,p} \sum_{q = 1}^n \left(\mathbf{\Lambda}_\mathbf{X}^{\phi,L} \right)_{p,q} \left( \mathbf{U}_\mathbf{X}^{\phi,L})^T \right)_{q,l} \\ & = \sum_{p=1}^n \left( \mathbf{U}_\mathbf{X}^{\phi,L} \right)_{k,p} \left(\mathbf{\Lambda}_\mathbf{X}^{\phi,L} \right)_{p,p} \left( \mathbf{U}_\mathbf{X}^{\phi,L}) \right)_{l,p} \\ & = \sum_{p=1}^n \left( \mathbf{U}_\mathbf{X}^{\phi,L} \right)_{k,p} \mathbb{E}_{h} \left(  (\hat{\mathbf{h}}_i^{\phi,L})_p (\hat{\mathbf{h}}_i^{\phi,L})_p | \mathbf{X} \right) \left( \mathbf{U}_\mathbf{X}^{\phi,L}) \right)_{l,p} \\ & = \mathbb{E}_{h} \left( \sum_{p=1}^n \left( \mathbf{U}_\mathbf{X}^{\phi,L} \right)_{k,p} (\hat{\mathbf{h}}_i^{\phi,L})_p (\hat{\mathbf{h}}_i^{\phi,L})_p \left( \mathbf{U}_\mathbf{X}^{\phi,L}) \right)_{l,p} | \mathbf{X} \right) \\ & = \mathbb{E}_{h} \left( \sum_{p=1}^n (\hat{\mathbf{h}}_i^{\phi,L})_p \left( \mathbf{U}_\mathbf{X}^{\phi,L}\right)^T_{p,k} \sum_{q=1}^n (\hat{\mathbf{h}}_i^{\phi,L})_q \left( \mathbf{U}_\mathbf{X}^{\phi,L})^T \right)_{q,l} | \mathbf{X} \right) \\ & = \mathbb{E}_{h} \left( \left(\hat{\mathbf{H}}_\mathbf{X}^{\phi,L} ( \mathbf{U}_\mathbf{X}^{\phi,L})^T \right)_{i,k} \left(\hat{\mathbf{H}}_\mathbf{X}^{\phi,L} ( \mathbf{U}_\mathbf{X}^{\phi,L})^T \right)_{i,l} | \mathbf{X} \right)
    \end{split}
\end{equation}
where we have used successively the definition of the conjugate kernel, its eigendecomposition, and that $\hat{\mathbf{h}}_i^{\phi,L} \sim \mathcal{N} (0, \mathbf{\Lambda}_\mathbf{X}^{\phi,L})$. By the law of total expectations, they also have the same covariance structure with respect to $(\mathbf{X},h)$.

Note that this was also the case in the linear case (with the same calculation), and it was precisely the covariance structure that gave us that the random matrices have the same distribution, since a multivariate Gaussian is uniquely determined by its covariance matrix.

We can again get rid of $ \mathbf{U}_\mathbf{X}^\phi$ in the last stage of the proof, since it corresponds to a mere basis change which does not change the spectral distribution. 

Now, we will utilise a recent result in random matrix theory \cite{mervelede2}, which states that the empirical spectral distribution of a symmetric matrix whose entries are measurable functions of independent random variables converges in distribution to that of a Gaussian symmetric matrix with the same covariance structure. In our case, since the random matrices $\hat{\mathbf{H}}_\mathbf{X}^{\phi,L} ( \mathbf{U}_\mathbf{X}^{\phi,L})^T$ and $\phi( \mathbf{H}_\mathbf{X}^{L-1})$  have the same covariance structure, then the asscociated Gram matrices $K_{\mathbf{X},\mathbf{X}}^{\phi,L,N} = \frac{1}{N} \phi (\mathbf{H}_\mathbf{X}^{L-1})^T \phi (\mathbf{H}_\mathbf{X}^{L-1})$ and $ \frac{1}{N}  \mathbf{U}_\mathbf{X}^{\phi,L} \left(\hat{\mathbf{H}}_\mathbf{X}^{\phi,L} \right)^T \hat{\mathbf{H}}_\mathbf{X}^{\phi,L} ( \mathbf{U}_\mathbf{X}^{\phi,L})^T$ have the same limiting spectral distribution (if it exists), by Theorems 1 and 5 in \cite{mervelede2}.

Let us first verify that the assumptions in Theorems 1 and 5 in \cite{mervelede2} hold true in our case. Again, we reason conditionally on $\mathbf{X}$. By conditioning on $\mathbf{X}$, we will show that the kernel random matrices of interest have the same limiting spectral distribution for any infinite realisation of $\mathbf{X}$, i.e.\ when the randomness only comes from $h$.

To stick to the way they define random matrices, namely as matrices extracted from infinite random fields, we need to consider from the beginning an infinite realisation $K^{\phi,L-1}_{\mathbf{X},\mathbf{X}} \in \mathbb{R}^{\mathbb{N}^2}$, which therefore depends on $\psi$. To re-utilise their notations, consider $(\xi_{i,j})_{(i,j) \in \mathbb{Z}^2}$ the random field defined by i.i.d.\ samples of $\mathcal{N} (0,1)$.
We define a function $g: \mathbb{R}^{\mathbb{Z}^2} \rightarrow \mathbb{R}$, again to re-utilise their notation, as follows: 
$g$ only acts on a portion of $\mathbb{R}^{\mathbb{Z}^2}$, namely $\mathbb{R}^{\mathbb{N}^2}$, and considering $ \mathbf{\Xi} = (\xi_{i,j})_{(i,j) \in \mathbb{N}^2}$, $g$ is then defined by:
\begin{equation}
    g \left( (\xi_{i,j})_{(i,j) \in \mathbb{Z}^2} \right) = \left( \phi \left(  \mathbf{\Xi} (\mathbf{\Lambda}_\mathbf{X}^{\phi,L-1})^{1/2} ( \mathbf{U}_\mathbf{X}^{\phi,L-1})^T \right) \right)_{0,0}
\end{equation}
where $\mathbf{\Lambda}_\mathbf{X}^{\phi,L-1}$ and$( \mathbf{U}_\mathbf{X}^{\phi,L-1})^T$is obtained by "diagonalisation" of the infinite matrix $K^{\phi,L-1}_{\mathbf{X},\mathbf{X}}$. 

Here, we are in a particular case where it is possible to define formally what is meant by "diagonalisation" of an infinite matrix. Indeed, the infinite matrix $K^{\phi,L-1}_{\mathbf{X},\mathbf{X}}$ can be constructed step by step: to go from size $n \times n$ to size $n+1 \times n+1$, we sample a new $\mathbf{x}_{n+1}$ from the teacher distribution. Let us note, just for this paragraph, simply $\mathbf{K}_n$ and $\mathbf{K}_{n+1}$ for these two intermediate matrices. The sub-matrix of size $n \times n$, extracted from $\mathbf{K}_{n+1}$, stays unchanged (it is equal to $\mathbf{K}_n$), as we simply add a row and a column, namely $\left(K^{\phi,L-1}_{\mathbf{X},\mathbf{x}_{n+1}}\right)^T$ and $K^{\phi,L-1}_{\mathbf{X},\mathbf{x}_{n+1}}$ with the same notation as equations \eqref{nngp mean}-\eqref{nngp variance}.

We decompose $\frac{1}{d} \mathbf{K}_{n+1}$ into two symmetric matrices $\mathbf{A}$ and $\mathbf{B}$ 
\begin{equation}
    \frac{1}{d} \mathbf{K}_{n+1} =  \begin{pmatrix}
    \mathlarger{\mathlarger{\mathlarger{\mathlarger{\frac{1}{d} \mathbf{K}_n}}}} & \begin{matrix}
        0 \\ \vdots \\ 0
    \end{matrix} \\ \begin{matrix}
        0 & \cdots & 0
    \end{matrix} & 0
    \end{pmatrix} + \begin{pmatrix}
    \mathlarger{\mathlarger{\mathlarger{\mathlarger{0}}}} & \begin{matrix}
        \frac{1}{d} K^{\phi,L-1} (\mathbf{x}_1, \mathbf{x}_{n+1}) \\ \vdots \\ \frac{1}{d} K^{\phi,L-1} (\mathbf{x}_n, \mathbf{x}_{n+1})
    \end{matrix} \\ \begin{matrix}
        \frac{1}{d} K^{\phi,L-1} (\mathbf{x}_1, \mathbf{x}_{n+1} & \cdots & \frac{1}{d} K^{\phi,L-1} (\mathbf{x}_n, \mathbf{x}_{n+1})
    \end{matrix} & \frac{1}{d} K^{\phi,L-1} (\mathbf{x}_{n+1}, \mathbf{x}_{n+1})
    \end{pmatrix}
\end{equation}

$\mathbf{A}$ and $\mathbf{B}$ are almost commuting with respect to the Frobenius norm.
\begin{equation}
    \mathbf{A}\mathbf{B}= \begin{pmatrix}
    \mathlarger{\mathlarger{\mathlarger{\mathlarger{0}}}} & \begin{matrix}
        \frac{1}{d^2} \sum_{i=1}^n K^{\phi,L-1} (\mathbf{x}_1, \mathbf{x}_i) K^{\phi,L-1} (\mathbf{x}_i, \mathbf{x}_{n+1}) \\ \vdots \\ \frac{1}{d^2} \sum_{i=1}^n K^{\phi,L-1} (\mathbf{x}_n, \mathbf{x}_i) K^{\phi,L-1} (\mathbf{x}_i, \mathbf{x}_{n+1})
    \end{matrix}
    \end{pmatrix}
\end{equation}
and 
\begin{equation}
    \mathbf{B}\mathbf{A}= \begin{pmatrix}
    \mathlarger{\mathlarger{\mathlarger{\mathlarger{0}}}} \\ \begin{matrix}
        \frac{1}{d^2} \sum_{i=1}^n K^{\phi,L-1} (\mathbf{x}_1, \mathbf{x}_i) K^{\phi,L-1} (\mathbf{x}_i, \mathbf{x}_{n+1}) & \cdots & \frac{1}{d^2} \sum_{i=1}^n K^{\phi,L-1} (\mathbf{x}_n, \mathbf{x}_i) K^{\phi,L-1} (\mathbf{x}_i, \mathbf{x}_{n+1})
    \end{matrix}
    \end{pmatrix}
\end{equation}
One can verify that the Frobenius norm of $\mathbf{A}\mathbf{B}- \mathbf{B}\mathbf{A}$ is with high probability of the order $\mathcal{O}_{n,d \to \infty} ( \frac{n^{3/2}}{d^{2}})$, which is $o_{d \to \infty} (1)$ (i.e.\ $\to 0$), using that $K^{\phi,L-1} (\mathbf{x},\mathbf{x}') = \mathcal{O}_{d \to \infty} (1)$ (i.e.\ is bounded as $d \to \infty$) with high probability.

Thus, by \cite{almostcommuting}, there exists two matrices $\mathbf{A}'$ and $\mathbf{B}'$, which can be arbitrarily close to $\mathbf{A}$ and $\mathbf{B}$ (with respect to the operator norm), such that $\mathbf{A}'$ and $\mathbf{B}'$ commute. 

$\mathbf{A}'$ and $\mathbf{B}'$ commute, thus they are co-diagonalisable, i.e.\ they share an eigenvector basis, which allows for their spectra to be summed. By continuity of the spectrum of a matrix with respect to the operator norm, it suffices to study the spectrum of $\mathbf{A}' + \mathbf{B}'$ in lieu of $\mathbf{A}+ \mathbf{B}$. This ensures that the $n+1$ eigenvalues and the $n+1$ eigenvectors of $\frac{1}{d} \mathbf{K}_{n+1}$ can be obtained from the $n$ eigenvalues and eigenvectors of $\frac{1}{d} \mathbf{K}_{n}$ by adding one eigenvalue and concatenating one dimension to the eigenvector at each step. It is thus possible to define sequences of eigenvalues and eigenvectors, respectively $\left(\lambda_n^{\mathbf{X},\phi,L-1}\right)_{n \in \mathbb{N}}$ and $\left(u_n^{\mathbf{X},\phi,L-1}\right)_{n \in \mathbb{N}}$ to characterize a realisation of the infinite random matrix $K^{\phi,L-1}_{\mathbf{X},\mathbf{X}}$. Here, the infinite eigenvectors are given by $(u_n^{\mathbf{X},\phi,L-1})_i = \Phi_n (\mathbf{x}_i)$ with $\Phi_n$ the Mercer's eigenfunctions of kernel $K^{\phi,L-1}$. With these sequences, the entries of the infinite matrix $\Tilde{\mathbf{H}}_\mathbf{X}^{L-1} :=  \mathbf{\Xi} (\mathbf{\Lambda}_\mathbf{X}^{\phi,L-1})^{1/2} ( \mathbf{U}_\mathbf{X}^{\phi,L-1})^T$ are well-defined:
\begin{equation}
    \left(\Tilde{\mathbf{H}}_\mathbf{X}^{L-1}\right)_{i,j} = \sum_{k=0}^\infty \xi_{i,k} \; \sqrt{\lambda_k^{\mathbf{X},\phi,L-1}} \left(u_k^{\mathbf{X},\phi,L-1}\right)_j
\end{equation}
  
$ \mathbf{\Xi} (\mathbf{\Lambda}_\mathbf{X}^{\phi,L-1})^{1/2} ( \mathbf{U}_\mathbf{X}^{\phi,L-1})^T$ is precisely designed to reproduce realisations of the random matrix $\phi \left(\mathbf{H}_\mathbf{X}^{L-1}\right)$ (whose rows are independently sampled from $\mathcal{GP} (0, K^{\phi,L-1})$) with an underlying array of i.i.d.\ random variables $ \mathbf{\Xi}$. One can indeed verify that we have:
\begin{equation}
    \mathbb{E}_{\xi} \left( \left(\Tilde{\mathbf{H}}_\mathbf{X}^{L-1}\right)_{i,k} \left(\Tilde{\mathbf{H}}_\mathbf{X}^{L-1}\right)_{i,l} \right) =  K^{\phi,L-1} (\mathbf{x}_k, \mathbf{x}_l) 
\end{equation} which holds $\forall \;i, k, l \in \mathbb{N}$.

One can then verify that the careful indexing yields:
\begin{equation}
    g \left( (\xi_{k-i,l-j})_{(i,j) \in \mathbb{Z}^2} \right) = \left(\phi \left( \mathbf{H}_\mathbf{X}^{L-1} \right)\right)_{k,l}
\end{equation}
Note that we are still reasoning conditionally on $\mathbf{X}$ so we should have written $g_\mathbf{X}$ instead of $g$.
So far, we thus have that:
\begin{equation}
    \left(\phi \left( \mathbf{H}_\mathbf{X}^{L-1} \right)\right)_{k,l} \;| \mathbf{X} \sim g \left( \xi_{k-i,l-j})_{(i,j) \in \mathbb{Z}^2} \right)
\end{equation}

Furthermore, $g$ is a measurable function from $\mathbb{R}^{\mathbb{Z}^2}$ to $\mathbb{R}$, as composition, product and countable sum of measurable functions \cite{measure} ($\phi$ is measurable by assumption of the present theorem). We have reproduced the random field $\phi \left( (\mathbf{H}_\mathbf{X}^{L-1} \right)$ as a measurable function of a random field of i.i.d\ variables, thus the Theorem 5 in \cite{mervelede2} is applicable and we can study the limiting spectral distribution of the Gram matrix associated with a Gaussian matrix with the same covariance structure, in lieu of the spectral distribution of the associated Gram matrix $K_{\mathbf{X},\mathbf{X}}^{\phi,L,N}$. 

We also have to prove that the assumptions of Theorem 1 in \cite{mervelede2} also hold true for $\hat{\mathbf{H}}_\mathbf{X}^{\phi,L}$, to conclude that they have the same limiting spectral distribution (the common limiting spectral distribution of the Gram matrices of their corresponding Gaussian matrix). The exact same technique can be used to define $g$ in a similar fashion, using the "diagonalisation" of $K^{\phi,L-1}_{\mathbf{X},\mathbf{X}}$.

There is then the final stage of relaxing the conditioning on $\mathbf{X}$ to conclude that the random matrices $K_{\mathbf{X},\mathbf{X}}^{\phi,L,N}$ and $ \frac{1}{N}  \mathbf{U}_\mathbf{X}^{\phi,L} \left(\hat{\mathbf{H}}_\mathbf{X}^{\phi,L} \right)^T \hat{\mathbf{H}}_\mathbf{X}^{\phi,L} ( \mathbf{U}_\mathbf{X}^{\phi,L})^T$, whose randomness come from $(\mathbf{X}, h)$, have the same limiting spectral measure. It is allowed to do so because by conditioning on $\mathbf{X}$, we have showed that the kernel random matrices of interest have the same limiting spectral distribution \emph{for any infinite realisation of $\mathbf{X}$}, they thus have the same limiting spectral distribution without conditioning.

Thus, the proof can proceed as previously, since we have reduced the problem to studying the existence of the limiting spectral distribution of $\frac{1}{N} \left(\hat{\mathbf{H}}_\mathbf{X}^{\phi,L} \right)^T \hat{\mathbf{H}}_\mathbf{X}^{\phi,L}$. The limiting spectral distribution is therefore given by the Marchenko-Pastur map of the limiting spectral distribution of $\mathbf{\Lambda}_\mathbf{X}^{\phi,L}$ (if it exists). By assumption of the present theorem, it exists and it is the limiting spectral distribution $\mu_\psi^\phi$ of the conjugate kernel matrix $K_{\mathbf{X},\mathbf{X}}^{\phi,L}$, which terminates the proof. 

\section{Proof of Theorem 3.}

The generalization error of kernel regression with kernel $K^N$ is given by:
\begin{equation}
    E_{\mathcal{K}} (K^N, \mathbf{X}, \mathbf{y}, \mathbf{x}, y, \bar{z}_{K^N}) := y^2 - 2 y \; \bar{\mu}^{K^N}_{\mathbf{X},\mathbf{y},x} + (\bar{\mu}^{K^N}_{\mathbf{X},\mathbf{y},x})^2
\end{equation}
Similarly, the generalization error of Gaussian process regression with GP prior $z_{K^N} \sim \mathcal{GP} (0, K^N)$ is given by:
\begin{equation}
    E_{\mathcal{GP}} (K^N, \mathbf{X}, \mathbf{y}, \mathbf{x}, y, \bar{z}_{K^N}) := y^2 - 2 y \; \bar{z}_{K^N} + (\bar{z}_{K^N})^2
\end{equation}
Averaging over the prediction distribution $\bar{z}_{K^N}$ yields (see equations \eqref{nngp mean}-\eqref{nngp variance}):
\begin{equation}
    \begin{split}
        E_{\mathcal{GP}} (K^N, \mathbf{X}, \mathbf{y}, \mathbf{x}, y) & = y^2 - 2 y \; \bar{\mu}^{K^N}_{\mathbf{X},\mathbf{y},x} + (\bar{\mu}^{K^N}_{\mathbf{X},\mathbf{y},x})^2 + \bar{K}_{\mathbf{X},\mathbf{y},x}^N \\ & = E_{\mathcal{K}} (K^N, \mathbf{X}, \mathbf{y}, \mathbf{x}, y) + \bar{K}_{\mathbf{X},\mathbf{y},x}^N
    \end{split}
\end{equation}
We will first study $E_{\mathcal{K}} (K^N, \mathbf{X}, \mathbf{y}, \mathbf{x}, y)$, and then $\bar{K}_{\mathbf{X},\mathbf{y},x}^N$.

Conditionally on $\mathbf{x}$, $y$, $\mathbf{X}$, $\mathbf{y}$, we expand $E_{\mathcal{K}} (K^N, \mathbf{X}, \mathbf{y}, \mathbf{x}, y)$ using the formula for the prediction mean of Gaussian process regression (see equation \eqref{nngp mean}:
\begin{equation}
    \begin{split}
    E_{\mathcal{K}} (K^N, \mathbf{X}, \mathbf{y}, \mathbf{x}, y) := & y^2 - 2 y (K_{\mathbf{x},\mathbf{X}}^N)^T \left(K_{\mathbf{X},\mathbf{X}}^N + \sigma_\epsilon^2 \mathbf{I}_n\right)^{-1} \mathbf{y}
    \\ & + (K_{\mathbf{x},\mathbf{X}}^N)^T \left(K_{\mathbf{X},\mathbf{X}}^N + \sigma_\epsilon^2 \mathbf{I}_n\right)^{-1} \mathbf{y} \mathbf{y}^T \left(K_{\mathbf{X},\mathbf{X}}^N + \sigma_\epsilon^2 \mathbf{I}_n\right)^{-1} K_{\mathbf{x},\mathbf{X}}^N 
    \end{split}
\end{equation}

Averaging over the out-of-sample data $\mathbf{x}$ and $y$, this yields:
\begin{equation}
    \begin{split}
    E_{\mathcal{K}} (K^N, \mathbf{X}, \mathbf{y}) &:= \mathbb{E}_{\mathbf{x},y} \left(E_{\mathcal{K}} (K^N, \mathbf{X}, \mathbf{y}, \mathbf{x}, y) \right) \\ & =\mathbb{E}_x\left( f_d(\mathbf{x})^2\right) + \sigma_\tau^2 - 2 \; \mathrm{Tr} \left(   \mathbb{E}_x \left( f_d(\mathbf{x}) K_{\mathbf{x},\mathbf{X}}^N \right) \; \mathbf{y}^T \left(K_{\mathbf{X},\mathbf{X}}^N + \sigma_\epsilon^2 \mathbf{I}_n\right)^{-1} \right) \\ & \; \; \; +  \mathrm{Tr} \left( \mathbb{E}_x \left(K_{\mathbf{x},\mathbf{X}}^N (K_{\mathbf{x},\mathbf{X}}^N)^T \right) \; \left(K_{\mathbf{X},\mathbf{X}}^N  + \sigma_\epsilon^2 \mathbf{I}_n\right)^{-1} \mathbf{y} \mathbf{y}^T \left(K_{\mathbf{X},\mathbf{X}}^N  + \sigma_\epsilon^2 \mathbf{I}_n \right)^{-1}  \right)
    \end{split}
\end{equation}
where we have used the property $\mathbf{a}^T \mathbf{b} = \mathrm{Tr} (\mathbf{a} \mathbf{b}^T)$. 

We use the Mercer's decomposition \cite{mercertheorem} of the random kernel $K^N$, which gives us:
\begin{equation}
    K_{\mathbf{X},\mathbf{X}}^N =  \mathbf{\Phi} \mathbf{\Lambda}  \mathbf{\Phi}^T
\end{equation}
with $\mathbf{\Lambda} = \mathrm{diag} (\lambda_1, ...., \lambda_M)$ the Mercer's eigenvalues, and $ \mathbf{\Phi}_{i,j} = \Phi_j (\mathbf{x}_i)$ the Mercer's eigenfunctions evaluated at training examples. Note that $\mathbf{\Lambda} \in \mathbb{R}^{M \times M}$, and $ \mathbf{\Phi} \in \mathbb{R}^{n \times M}$ with $\frac{n}{M} \to 0$ as $n, M \to \infty$ (i.e.\ $M$ going to infinity at a much faster rate than $n$). More importantly, note that here, kernel eigenvalues $\lambda_i$ do depend on the data distribution $\mathbb{P}_d$ by Mercer's theorem, but not on the particular realisation of the training set $\mathbf{X}$. The randomness of $\lambda_i$ only comes from $K^N$.
We then have:
\begin{equation}\label{conditonedgeneralisation}
    \begin{split}
        E_{\mathcal{K}} (K^N, \mathbf{X}, \mathbf{y}) = & \mathbb{E}_x\left( f_d(\mathbf{x})^2\right) + \sigma_\tau^2- \mathbb{E}_x \left(f_d(\mathbf{x}) \sum_i^{M} \frac{1}{\lambda_i + \sigma_\epsilon^2} \left( \mathbf{\Phi}^\dagger \mathbf{y} (\mathbf{k}^N_{\mathbf{x},\mathbf{X}})^T ( \mathbf{\Phi}^\dagger)^T \right)_{i,i} \right) \\ & + \sum_{i}^{M} \frac{1}{(\lambda_i + \sigma_\epsilon^2)^2} \left( \mathbf{\Phi}^\dagger \mathbf{y} \mathbf{y}^T ( \mathbf{\Phi}^\dagger)^T \right)_{i,i} \mathbb{E}_x \left(\left( \mathbf{\Phi}^\dagger \mathbf{k}^N_{\mathbf{x},\mathbf{X}}  (\mathbf{k}^N_{\mathbf{x},\mathbf{X}})^T ( \mathbf{\Phi}^\dagger)^T \right)_{i,i} \right) \\ & + \sum_{i}^{M} \sum_{j \neq i}^{M}\frac{1}{(\lambda_i + \sigma_\epsilon^2)(\lambda_j + \sigma_\epsilon^2)} \left( \mathbf{\Phi}^\dagger \mathbf{y} \mathbf{y}^T ( \mathbf{\Phi}^\dagger)^T \right)_{i,j} \mathbb{E}_x \left(\left( \mathbf{\Phi}^\dagger \mathbf{k}^N_{\mathbf{x},\mathbf{X}}  (\mathbf{k}^N_{\mathbf{x},\mathbf{X}})^T ( \mathbf{\Phi}^\dagger)^T \right)_{i,j} \right)
    \end{split}
\end{equation} 
where $  \mathbf{\Phi}^\dagger =  \mathbf{\Phi}^T ( \mathbf{\Phi}  \mathbf{\Phi}^T)^{-1} $ is the Moore-Penrose pseudo-inverse \cite{moorepenrose} of $ \mathbf{\Phi}$, which can be calculated this way because $ \mathbf{\Phi}$ has orthogonal rows. 

To relax the conditioning on $K^N$, our objective is to utilise that, in the limit of infinite quantities $N, n, d \to \infty$, the expectations of $\frac{1}{\lambda_i + \sigma_\epsilon^2}$ and $\frac{1}{(\lambda_i + \sigma_\epsilon^2)^2}$, with respect to $K^N$, are easy to calculate: they are respectively given by  $g (\gamma, \psi)$ and $g_2 (\gamma, \psi)$ (see equation \eqref{g function}). But this cannot be used directly because in the three sums in equation (\ref{conditonedgeneralisation}), the $\frac{1}{\lambda_i }$ and  $\frac{1}{\lambda_j}$ are multiplied by functions of $\mathbf{k}^N_{\mathbf{x},\mathbf{X}}$ which also have a dependency on $\lambda_i, \lambda_j$.

To tackle this issue, we rewrite $\mathbf{k}^N_{\mathbf{x},\mathbf{X}} =  \mathbf{\Phi} \mathbf{\Lambda}  \mathbf{\Phi}^*$ with $ \mathbf{\Phi}^*_j = \Phi_j (\mathbf{x})$ the eigenfunction evaluated at the test point $\mathbf{x}$ and we decompose the different quantities:
\begin{equation}
    \begin{split}
    \left( \mathbf{\Phi}^\dagger \mathbf{y} (\mathbf{k}^N_{\mathbf{x},\mathbf{X}})^T ( \mathbf{\Phi}^\dagger)^T \right)_{i,i} & = \sum_{k,l} \left( \mathbf{\Phi}^\dagger \mathbf{y} ( \mathbf{\Phi}^*)^T\right)_{i,k} \mathbf{\Lambda}_{k,l} \left( \mathbf{\Phi}^T ( \mathbf{\Phi}^\dagger)^T\right)_{l,i} \\& = \sum_{k} \lambda_k \left( \mathbf{\Phi}^\dagger \mathbf{y} ( \mathbf{\Phi}^*)^T\right)_{i,k}  \left( \mathbf{\Phi}^\dagger  \mathbf{\Phi} \right)_{i,k}
    \end{split}
\end{equation}
The first sum can therefore be separated into two parts:
\begin{equation}
    \begin{split}
    \sum_i^{M} \frac{1}{\lambda_i + \sigma_\epsilon^2} \left( \mathbf{\Phi}^\dagger \mathbf{y} (\mathbf{k}^N_{\mathbf{x},\mathbf{X}})^T ( \mathbf{\Phi}^\dagger)^T \right)_{i,i} = & \sum_i^{M} \frac{\lambda_i}{\lambda_i + \sigma_\epsilon^2} \left( \mathbf{\Phi}^\dagger \mathbf{y} ( \mathbf{\Phi}^*)^T\right)_{i,i}  \left( \mathbf{\Phi}^\dagger  \mathbf{\Phi} \right)_{i,i} \\& + \sum_i^{M} \frac{1}{\lambda_i + \sigma_\epsilon^2}\sum_{k \neq i} \lambda_k \left( \mathbf{\Phi}^\dagger \mathbf{y} ( \mathbf{\Phi}^*)^T\right)_{i,k}  \left( \mathbf{\Phi}^\dagger  \mathbf{\Phi} \right)_{i,k}
    \end{split}
\end{equation}
We have thus isolated the term $\frac{1}{\lambda_i + \sigma_\epsilon^2}$ in the second part which allows us to calculate the sum using the integral $\int_0^\infty \frac{1}{\lambda + \sigma_\epsilon^2} \mathrm{d} (\rho_{MP}^{\gamma} \boxtimes \mu_{\psi}^{\phi}) (\lambda)$ when averaging over $K^N$.

The same trick can be used to decompose the other terms in the sums:
\begin{equation}
    \begin{split}
    \left( \mathbf{\Phi}^\dagger \mathbf{k}^N_{\mathbf{x},\mathbf{X}} (\mathbf{k}^N_{\mathbf{x},\mathbf{X}})^T ( \mathbf{\Phi}^\dagger)^T \right)_{i,j} = \sum_{k_1,k_2} \lambda_{k_1} \lambda_{k_2} \left( \mathbf{\Phi}^* ( \mathbf{\Phi}^*)^T\right)_{k_1,k_2} \left( \mathbf{\Phi}^\dagger  \mathbf{\Phi} \right)_{i,k_1}  \left( \mathbf{\Phi}^\dagger  \mathbf{\Phi} \right)_{j,k_2}
    \end{split}
\end{equation}
which allows us to isolate the relevant terms and integrate the spectral measures when averaging over $K^N$. This gives us the expression in equation \eqref{kernel reg error eq}. 

The same techniques can be used to study $\bar{K}_{\mathbf{X},\mathbf{y},x}^N$. We obtain:
\begin{equation}
    \mathbb{E}_x \left(\bar{K}_{\mathbf{X},\mathbf{y},x}^N\right) = \mathbb{E}_x \left(K^N (\mathbf{x},\mathbf{x})\right) - \sum_{i}^{M} \frac{1}{(\lambda_i + \sigma_\epsilon^2)} \mathbb{E}_x \left(\left( \mathbf{\Phi}^\dagger \mathbf{k}^N_{\mathbf{x},\mathbf{X}}  (\mathbf{k}^N_{\mathbf{x},\mathbf{X}})^T ( \mathbf{\Phi}^\dagger)^T \right)_{i,i} \right)
\end{equation}
where the average of $\frac{1}{(\lambda_i + \sigma_\epsilon^2)}$ can be isolated in the same way as previously, giving us the expression in equation \eqref{gp reg error eq}.

Lastly, it is tedious but rather straightforward to show that the quantities $\left( \mathbf{\Phi}^\dagger \mathbf{y} ( \mathbf{\Phi}^*)^T\right)_{i,k}  \left( \mathbf{\Phi}^\dagger  \mathbf{\Phi} \right)_{i,k}$ and $\left( \mathbf{\Phi}^* ( \mathbf{\Phi}^*)^T\right)_{k_1,k_2} \left( \mathbf{\Phi}^\dagger  \mathbf{\Phi} \right)_{i,k_1}  \left( \mathbf{\Phi}^\dagger  \mathbf{\Phi} \right)_{j,k_2}$ are bounded when $n,d,N \to \infty$, and, in particular, when $i=j$, that the expectation is non-zero. One way to calculate it is to develop the terms further to deal with sums over all entries of matrices $ \mathbf{\Phi}$ and $ \mathbf{\Phi}^\dagger$ and invoke that $( \mathbf{\Phi}  \mathbf{\Phi}^T)^{-1}$ follows the well-known inverse Wishart distribution \cite{Livan_2018,El_Karoui_2010}.

\section{Proofs of Corollary 1.}


Note that the assumption of the noise-free limit $\sigma_\epsilon \to 0$ is motivated by the fact that it is known \cite{gpforml} that the predicted mean of Gaussian process regression is the same as the prediction of kernel ridge regression with the same kernel $K$ and a ridge parameter of ${\sigma_\epsilon}^2$, hence a noise model is some form of ridge regularisation, when we actually want to recover the double descent in the ridgeless case, in order to have a proper divergence at the interpolation threshold. 

In the noise-free limit, since the kernel random matrix $\mathbf{K}^N_{\mathbf{X},\mathbf{X}} \in \mathbb{R}^{n \times n}$ is not necessarily invertible (it is at most of rank $\min (n,N)$ which is $\leq n$), we need to consider a generalized inversion in the formula for the prediction meaan of Gaussian process regression (equation \eqref{nngp mean}), namely the Moore-Penrose pseudo-inverse (\cite{moorepenrose}). In this case, we integrate only over the strictly positive part of the spectral measures in equation \eqref{kernel reg error eq}. This stems from the fact that the eigendecomposition of the Moore-Penrose pseudo-inverse of a matrix with eigendecomposition $\mathbf{V} \mathbf{\Lambda} \mathbf{V}^T$ is obtained by transposing $\mathbf{V}$ and $\mathbf{V}^T$, inverting the non-zero eigenvalues, and leaving in place the zero eigenvalues.

We next derive the behavior, with respect to $\gamma$, of the limiting generalisation errors $E_{\mathcal{K}}(\gamma, \psi)$ and  $E_{\mathcal{GP}}(\gamma, \psi)$.

\subsection{Underparameterized regime}

In the limit $\gamma \to \infty$, the Marchenko-Pastur fixed-point equation (equation \eqref{marchenkopasturfixedpoint}) becomes $S(z) = 0$. The Stieltjes inversion formula tells us that the solution is a probability measure with density $0$ for all $x \neq 0$: it is the Dirac measure at $0$. Plugging this into equation \eqref{kernel reg error eq}, and integrating over the strictly positive part of the spectral measures (we are precisely in the case where the rank of $\mathbf{K}^N_{\mathbf{X},\mathbf{X}}$ is $< n$, since $\frac{n}{N} \to \infty$), we recover equation \eqref{under regime}.

\subsection{Overparameterized regime}

As per equation \eqref{mp converge dirac}, the spectral measure of interest is that of the \emph{actual} conjugate kernel matrix. Plugging this into equation \eqref{kernel reg error eq}, we recover equation \eqref{over regime}.

\subsection{Interpolation threshold} 

For the case $\gamma \to 1$, we will show that $\lambda \mapsto \frac{1}{\lambda^2}$ is not integrable with respect to $\rho_{MP}^{1} \boxtimes \mu_\psi^\phi$. The limits $\gamma \to 1^-$ and $\gamma \to 1^+$ yield the same result, simply by integrating only over the strictly positive part of $\rho_{MP}^{1} \boxtimes \mu_\psi^\phi$ in the case $\gamma = \frac{n}{N} \to 1^+$.  Indeed, in $B(\gamma, \psi) \frac{1}{\lambda^2} + A(\gamma, \psi) \frac{1}{\lambda}$ and $B(\gamma, \psi) \frac{1}{\lambda^2} + \bar{A} (\gamma, \psi) \frac{1}{\lambda}$, the terms in $\frac{1}{\lambda^2}$ dominate near $0$, and they are thus the ones that will give us the divergence.

First, we notice that $\lambda \mapsto \frac{1}{\lambda^2}$ is not integrable with respect to the simple Marchenko-Pastur distribution $\rho_{MP}^1$ (which is the limiting spectral measure of $\mathbf{M}^T\mathbf{M}$, with the same notations as in equation (\ref{rewrittingh}), with $\gamma =1$). Indeed, with the analytical formula for the Marchenko-Pastur distribution \cite{marchenkopastur}: 
\begin{equation}
    \rho_{MP}^{1} (\lambda) = \frac{1}{2\pi\lambda}\sqrt{\lambda(2 - \lambda)}
\end{equation}
we have that $ \frac{1}{\lambda^{1/2}} = \mathcal{O}_{\lambda \to 0} (\rho_{MP}^1 (\lambda))$, thus $ \frac{1}{\lambda^{5/2}} = \mathcal{O}_{\lambda \to 0} (\frac{1}{\lambda^2}\rho_{MP}^1 (\lambda))$, and $\frac{1}{\lambda^{5/2}}$ is not integrable at $0$ (convergence of Riemann integrals), thus $\frac{1}{\lambda^2}\rho_{MP}^1 (\lambda)$ is not integrable at $0$.

The assumption on the integrability of $\lambda \mapsto \frac{1}{\lambda}$ and $\lambda \mapsto \frac{1}{\lambda^2}$ with respect to the spectral measure $\mu_\psi^\phi$ implies that, in the limit of infinite quantities, $\mathbf{\Lambda}_\mathbf{X}$, with the same notations as in equation (\ref{rewrittingh}), is almost surely invertible. Indeed, if it had a non-zero probability of having a zero eigenvalue, in other words if we did not have $\mu_\psi^\phi (\lambda) = \mathrm{o}_{\lambda \to 0} (1)$, $\lambda \mapsto \frac{1}{\lambda}$ would not have been integrable at $0$.

Next, we show that the Marchenko-Pastur map of $\mu_\psi^\phi$ with $\gamma =1$ can only worsen the non-integrability at $0$. Precisely, we show that $ \rho_{MP}^1  (\lambda) = \mathcal{O}_{\lambda \to 0} \left(\left(\rho_{MP}^1 \boxtimes \mu_\psi^\phi \right) (\lambda) \right)$. To do so, using the same notations as in equation (\ref{rewrittingh}), we show that if, in the infinite limit, $\mathbf{M}^T\mathbf{M}$ has an arbitrarily small eigenvalue $\epsilon > 0$ with probability $p$, then $\mathbf{\Lambda}_\mathbf{X}^{1/2} \mathbf{M}^T\mathbf{M}\mathbf{\Lambda}_\mathbf{X}^{1/2}$ has, with probability $p$, a smaller eigenvalue (this way the density near zero of the spectral measure of $\mathbf{M}^T\mathbf{M}$ is dominated). Suppose that $\mathbf{M}^T\mathbf{M}$ has a small eigenvalue $\epsilon > 0$ with associated eigenvector $e$. Note $\lambda_{min}^\epsilon$ the smallest eigenvalue of $\mathbf{\Lambda}_\mathbf{X}^{1/2} \mathbf{M}^T\mathbf{M}\mathbf{\Lambda}_\mathbf{X}^{1/2}$, conditional on the existence of an eigenvalue $\epsilon$ for the random matrix $\mathbf{M}^T\mathbf{M}$. We have:
\begin{equation}
    \begin{split}
    \left( \mathbf{\Lambda}_\mathbf{X}^{-1/2} e \right)^T \mathbf{\Lambda}_\mathbf{X}^{1/2} \mathbf{M}^T\mathbf{M}\mathbf{\Lambda}_\mathbf{X}^{1/2} \left( \mathbf{\Lambda}_\mathbf{X}^{-1/2} e \right) & = \epsilon || e ||^2 = \epsilon \\ & \geq \lambda_{min}^\epsilon|| \mathbf{\Lambda}_\mathbf{X}^{-1/2} e ||^2
    \end{split}
\end{equation}
where we have used that $e$ is an orthonormal eigenvector of $\mathbf{M}^T\mathbf{M}$ with eigenvalue $\epsilon$, and the standard inequality $\forall \mathbf{x} \in \mathbb{R}^n\;:\; \mathbf{x}^T \mathbf{A} \mathbf{x} \geq \lambda^\mathbf{A}_{min} || \mathbf{x} ||^2 $, which holds for any real symmetric matrix $\mathbf{A}$. We can express the norm of the vector of interest, calculated in the eigenvector basis of $\mathbf{M}^T\mathbf{M}$:
\begin{equation}
    || \mathbf{\Lambda}_\mathbf{X}^{-1/2} e ||^2 = \sum_i^n \frac{1}{\lambda_i}
\end{equation}
with $\lambda_i$ the eigenvalues of $\mathbf{\Lambda}_\mathbf{X}$.  By the law of large numbers, we have: 
\begin{equation}
    \lim_{n \to \infty} \frac{1}{n} \sum_i^n \frac{1}{\lambda_i} = \int_{0}^\infty \frac{1}{\lambda} d (\mu_\psi^\phi) (\lambda)
\end{equation} almost surely, and the limit is finite, by assumption on $\mu_\psi^\phi$ (limiting spectral measure of $\mathbf{\Lambda}_\mathbf{X}$). Thus, we have $ \lim_{n \to \infty} ||\mathbf{\Lambda}_\mathbf{X}^{-1/2} e ||^2 = \infty$ almost surely. Note that we are still conditioned on $\mathbf{M}$, the ``almost surely'' statement refers to the randomness of $\mathbf{X}$. Therefore, in the limit of infinite quantities, we have almost surely:
\begin{equation}
     \epsilon \geq \frac{\epsilon}{|| \mathbf{\Lambda}_\mathbf{X}^{-1/2} e ||^2} \geq \lambda_{min}^\epsilon
\end{equation}
Relaxing the conditioning on the existence of a eigenvalue $\epsilon$ for $\mathbf{M}^T\mathbf{M}$, we have thus found, with probability $p$, a smaller eigenvalue $\lambda_{min}^\epsilon$ for $\mathbf{\Lambda}_\mathbf{X}^{1/2} \mathbf{M}^T\mathbf{M}\mathbf{\Lambda}_\mathbf{X}^{1/2}$, which gives us the result we wanted. Now, since the Marchenko-Pastur map (with $\gamma =1$) of $\mu_\psi^\phi$ dominates the Marchenko-Pastur distribution $\rho_{MP}^1 (\lambda)$ near zero, it is not integrable at zero either. This gives us the result:
\begin{equation}
    \begin{split}
    & \lim_{\gamma \to 1} E_{\mathcal{K}} (\gamma, \psi) = \infty \\& \lim_{\gamma \to 1} E_{\mathcal{GP}} (\gamma, \psi) = \infty
    \end{split}
\end{equation}

\section{Numerical experiments.}

Our results about generalization errors of kernel and GP regression, including the double-descent phenomenon, rely on the fact that the random width-dependent NNGP kernel $K^N$ can be written:
\begin{equation}\label{kernel identity}
    K^N (\mathbf{x}, \mathbf{x}') = \mathbf{\Phi} (\mathbf{x})^T \mathbf{\Lambda} \mathbf{\Phi} (\mathbf{x}')
\end{equation}
where $\mathbf{\Phi} (\mathbf{x}) := (\Phi_1 (\mathbf{x}), ..., \Phi_M (\mathbf{x})) $, $\mathbf{\Phi} (\mathbf{x}') := (\Phi_1 (\mathbf{x}'), ..., \Phi_M (\mathbf{x}'))$ the Mercer's eigenfunctions evaluated at points $\mathbf{x}, \mathbf{x}'$, and $\mathbf{\Lambda} = \mathrm{diag} (\lambda_1, ..., \lambda_M)$ with $\lambda_i$ sampled independently from $\rho_{MP}^{\gamma} \boxtimes \mu_{\psi}^{\phi}$. 

\begin{figure}[ht]
\vskip 0.2in
\centerline{\includegraphics[height=4.5cm]{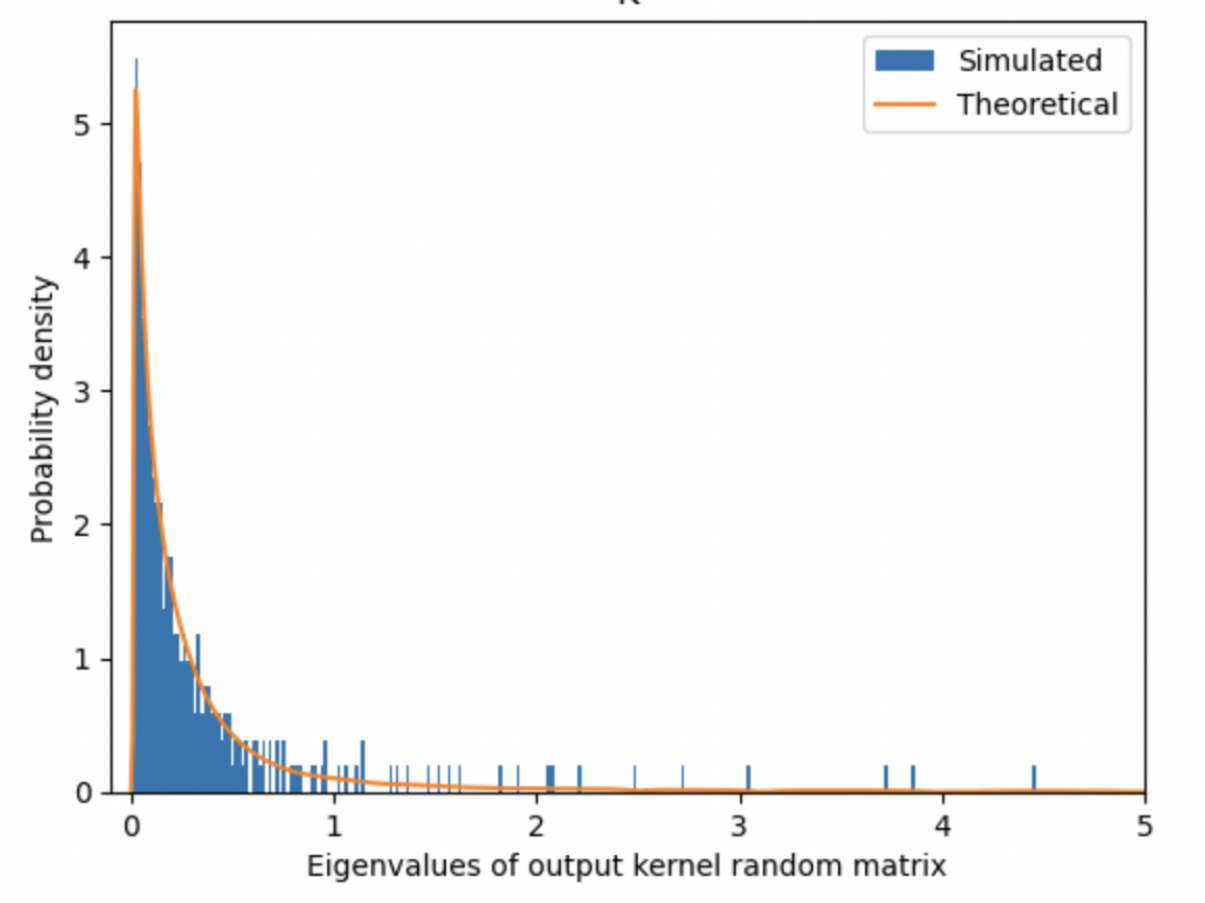}}
\caption{Simulated empirical spectral distribution versus theoretical limiting spectral distribution of the empirical covariance matrix $\mathbf{K}_{\mathbf{X},\mathbf{X}}^N$. We use a two-layer $\mathrm{tanh}$ NNGP on a subset of MNIST taking  $N = 600$, $n = 300$, and $d = 784$ (number of pixels on MNIST images).The simulated distribution is obtained by sampling from the random matrix, and the theoretical distribution is obtained by solving the Marchenko-Pastur fixed-point equation.}
\label{eigenvaluedistribtanh}
\end{figure}

In our numerical simulations, the spectral distribution of the actual NNGP kernel $\mu_{\psi}^{\phi}$ is estimated by diagonalising $\mathbf{K}^{\phi, \hat{N}}_{\mathbf{X},\mathbf{X}}$ with a value $\hat{N} \gg n$. The Marchenko-Pastur map is then estimated by solving the fixed-point equation (equation \eqref{marchenkopasturfixedpoint}) via iteration through the recursive sequence in the Stieljes transform space. Figure \ref{eigenvaluedistribtanh} complements the examples given in the main paper (Figure \ref{eigenvaluedistribsynth}) with the spectral distribution in the case of a two-layer (width-dependent) NNGP with $\mathrm{tanh}$ activation function on MNIST; our procedure provides a very good match. 

\begin{figure}[ht]
\vskip 0.2in
\begin{minipage}{.48\textwidth}
\centerline{\includegraphics[height=4.5cm]{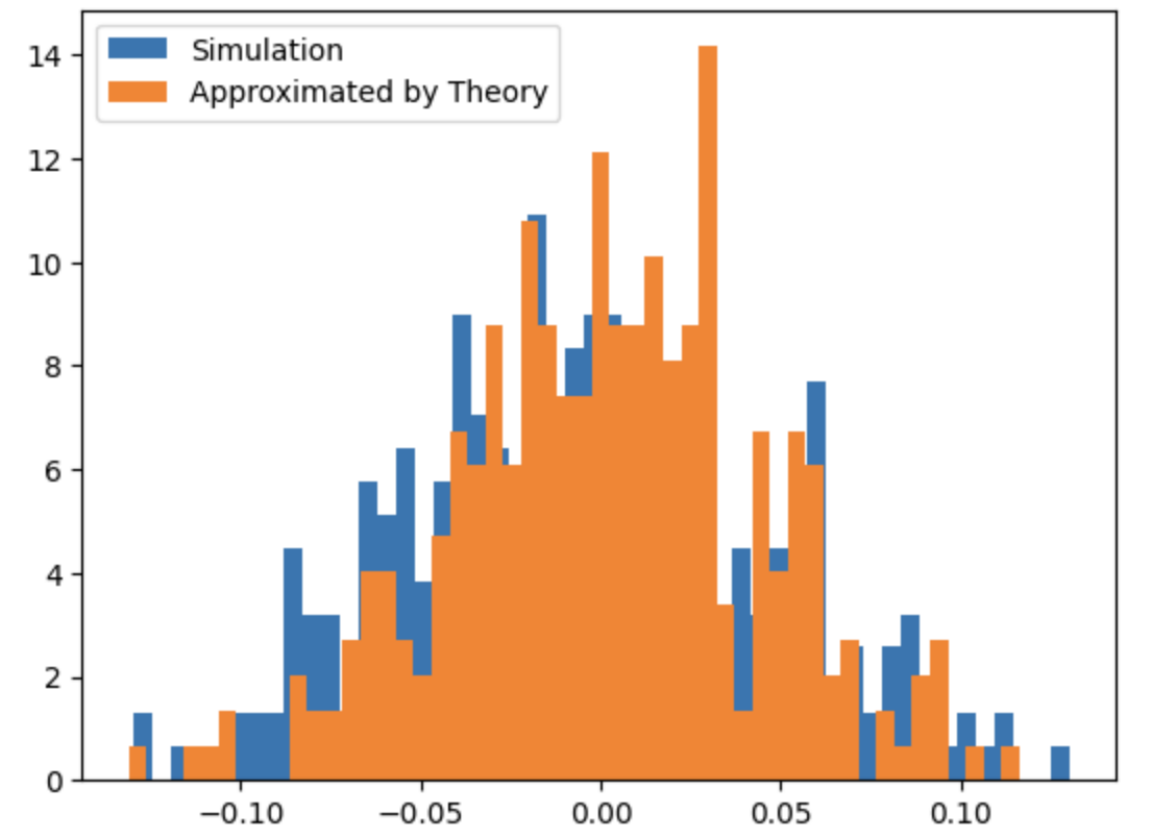}}
\end{minipage}
\begin{minipage}{.48\textwidth}
\centerline{\includegraphics[height=4.5cm]{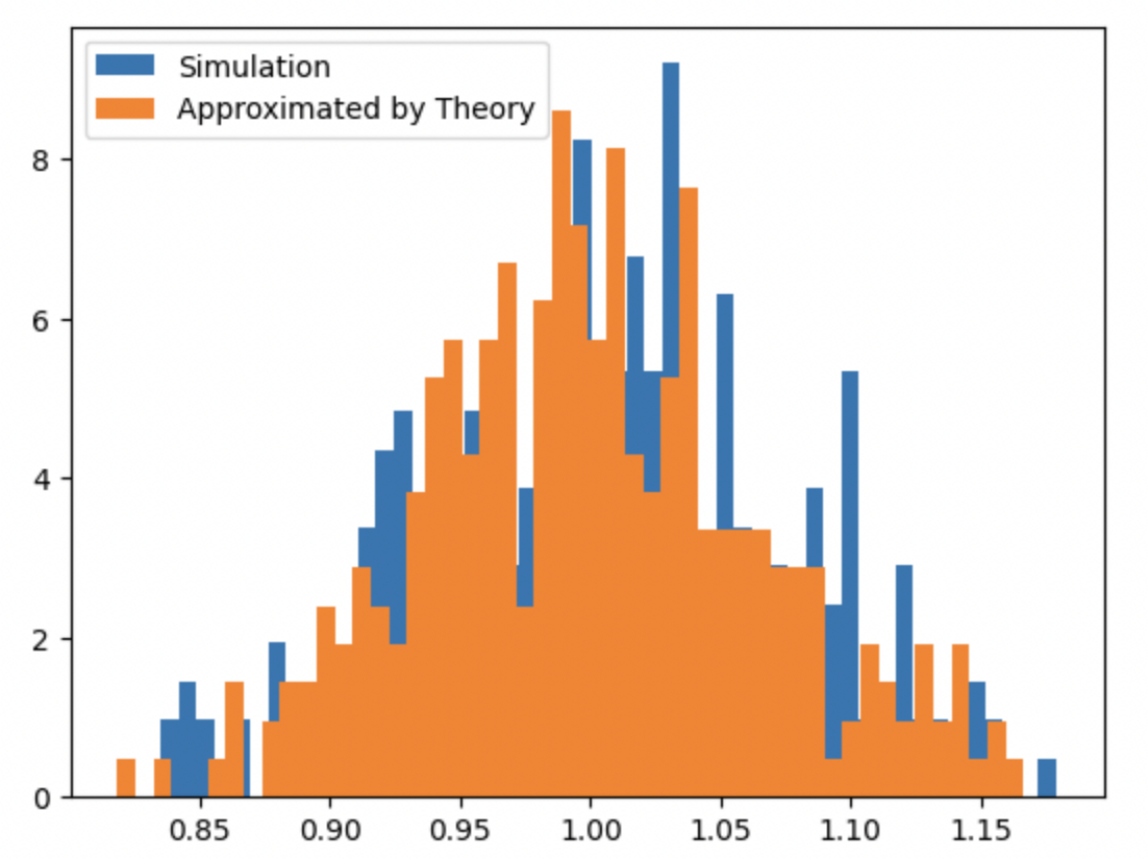}}
\end{minipage}
\caption{Distributions of random width-dependent NNGP kernel values. X-axis are indexed by the kernel value and Y-axis are indexed by the number of samples. On the left, the simulated kernel value is obtained by sampling from the off-diagonal elements of the kernel random matrix of a two-layer NNGP with no non-linearity $K^N (\mathbf{x}_i, \mathbf{x}_j) = \frac{1}{N} \mathbf{x}_i^T \mathbf{W}^T \mathbf{W} \mathbf{x}_j$ under teacher distribution $\mathcal{N} (0, \frac{1}{d} I_d)$, and the theoretical distribution is obtained by sampling from $\mathbf{\Phi} (\mathbf{x}_i)^T \mathbf{\Lambda} \mathbf{\Phi} (\mathbf{x}_j)$ using the spectral distribution $\rho_{MP}^{\gamma} \boxtimes  \rho_{MP}^{\psi}$ and sampling from independent Gaussian entries in lieu of the eigenfunctions (SUA). On the right, the simulated kernel value is obtained by the same procedure for on-diagonal elements $K^N (\mathbf{x}_i, \mathbf{x}_i)$.}
\label{kerneldistribisotrop}
\end{figure}

\begin{figure}[ht]
\vskip 0.2in
\begin{minipage}{.48\textwidth}
\centerline{\includegraphics[height=4.5cm]{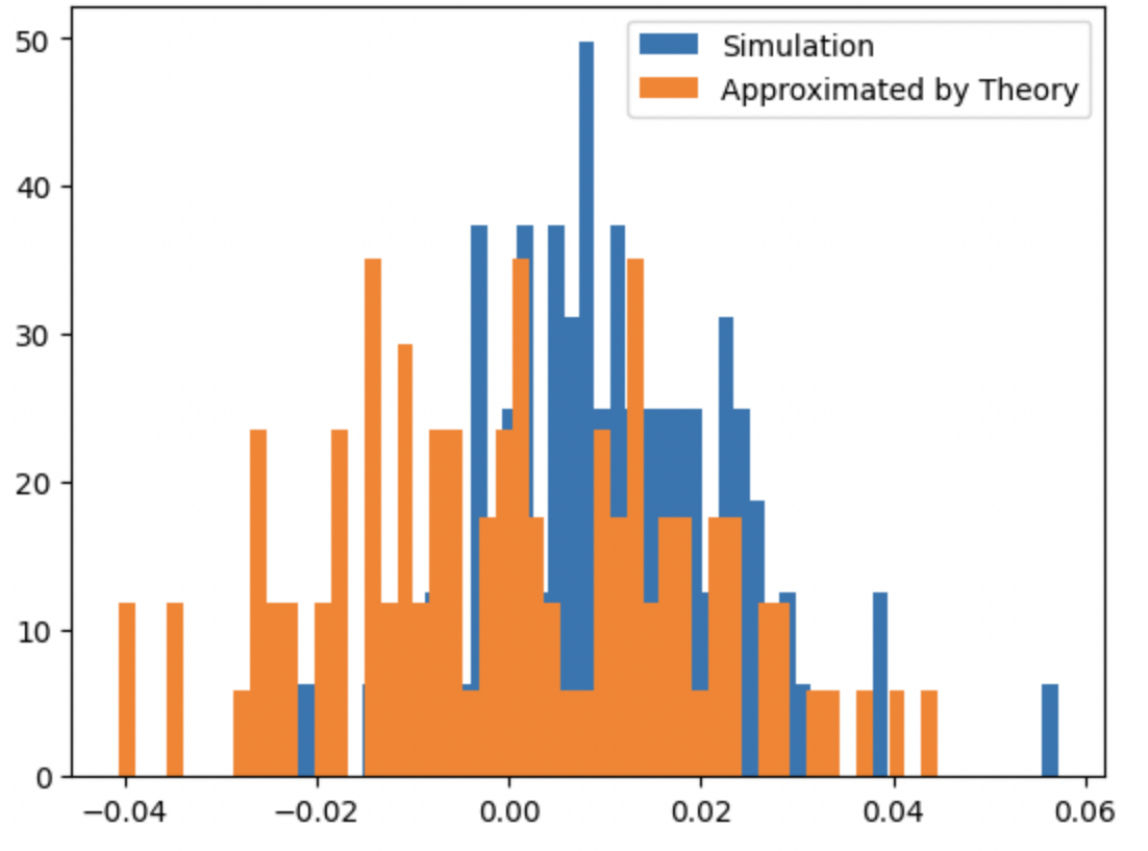}}
\end{minipage}
\begin{minipage}{.48\textwidth}
\centerline{\includegraphics[height=4.5cm]{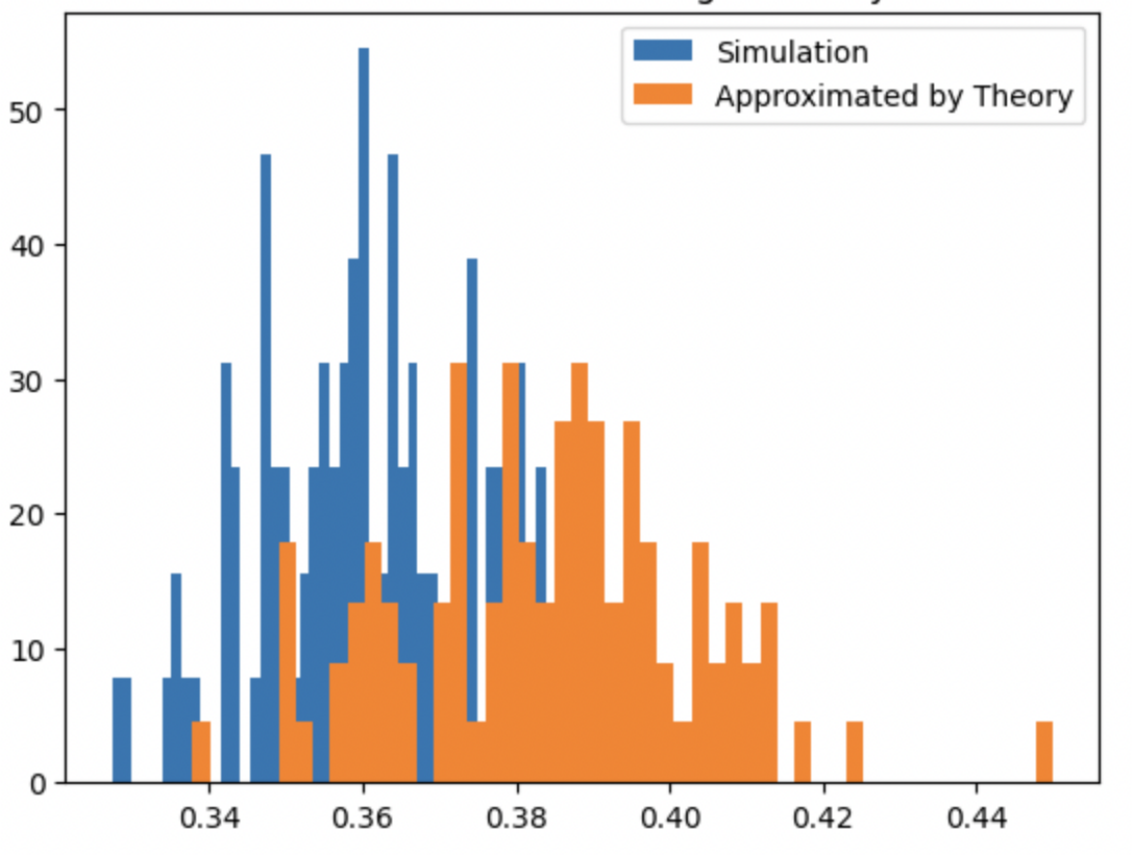}}
\end{minipage}
\caption{Distributions of random width-dependent NNGP kernel values. X-axis are indexed by the kernel value and Y-axis are indexed by the number of samples. On the left, the simulated kernel value is obtained by sampling from the off-diagonal elements of the kernel random matrix of a two-layer NNGP with $\phi = \mathrm{ReLU}$ nonlinearity $K^{\phi, N} (\mathbf{x}_i, \mathbf{x}_j) = \frac{1}{N} \phi \left(\mathbf{W} \mathbf{x}_i \right)^T \phi \left(\mathbf{W} \mathbf{x}_j \right)$ on a subset of MNIST, and the theoretical distribution is obtained by sampling from $\mathbf{\Phi} (\mathbf{x}_i)^T \mathbf{\Lambda} \mathbf{\Phi} (\mathbf{x}_j)$ using the spectral distribution of the actual NNGP kernel $\rho_{MP}^{\gamma} \boxtimes \mu_{\psi}^{\phi}$ and sampling from independent Gaussian entries in lieu of the eigenfunctions (SUA). On the right, the simulated kernel value is obtained by the same procedure for on-diagonal elements $K^{\phi,N} (\mathbf{x}_i, \mathbf{x}_i)$.}
\label{kerneldistribrelumnist}
\end{figure}

\begin{figure}[ht]
\vskip 0.2in
\begin{minipage}{.48\textwidth}
\centerline{\includegraphics[height=4.5cm]{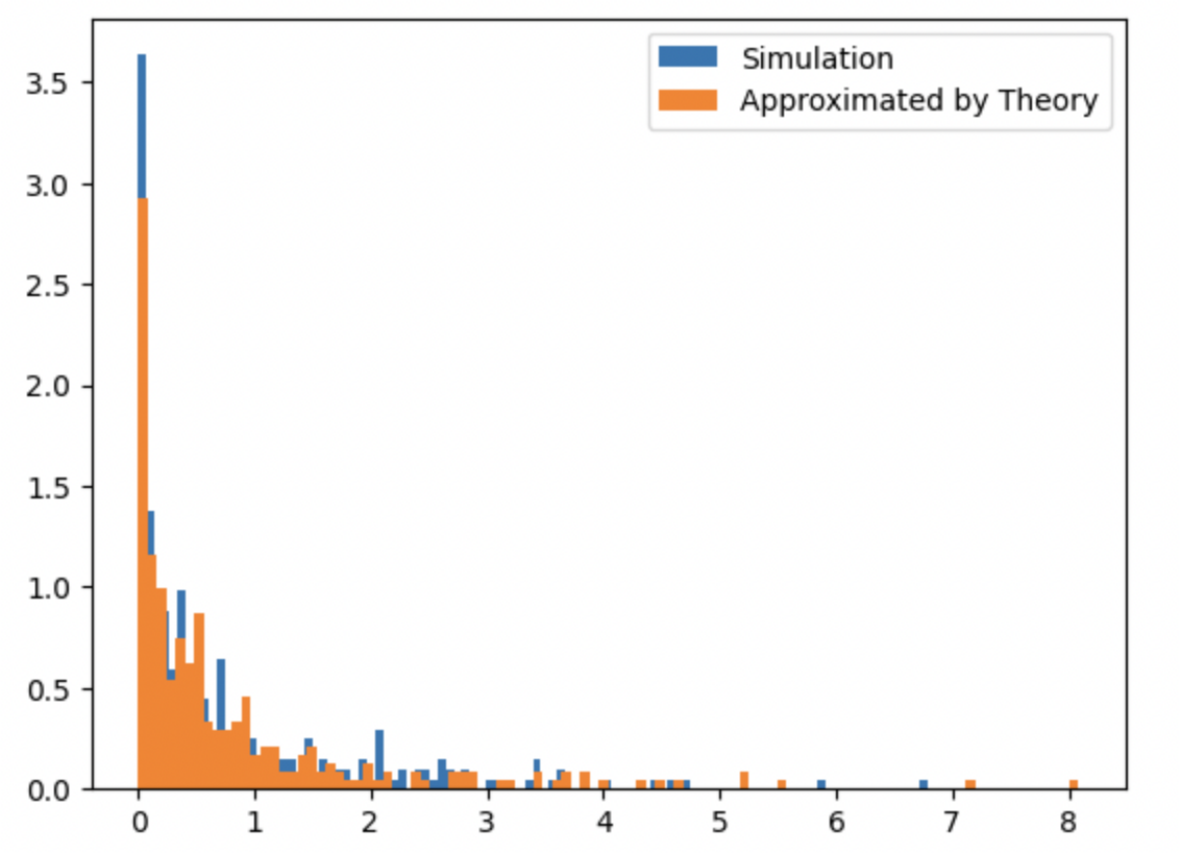}}
\end{minipage}
\begin{minipage}{.48\textwidth}
\centerline{\includegraphics[height=4.5cm]{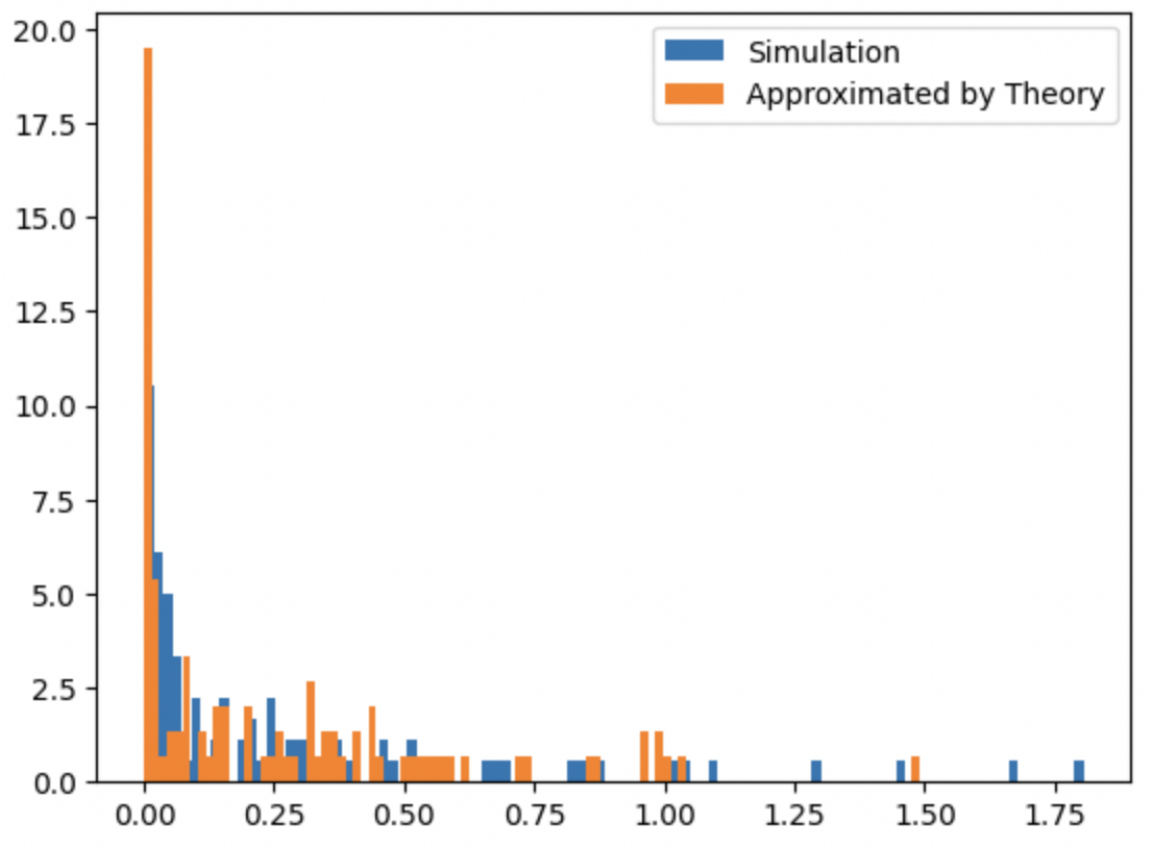}}
\end{minipage}
\caption{Distributions of generalisation errors of (random) kernel regression. X-axis are indexed by the generalization error and Y-axis are indexed by the number of samples. On the left, the simulated error is obtained by sampling from the prediction of kernel regression with the random (width-dependent) NNGP kernel $(\mathbf{x}, \mathbf{x}') \mapsto K^N (\mathbf{x}, \mathbf{x}')$ with no nonlinearity under teacher distribution $\mathcal{N} (0, \frac{1}{d} I_d)$, and the theoretical error is obtained by sampling from the approximate random kernel $(\mathbf{x}, \mathbf{x}') \mapsto \mathbf{\Phi} (\mathbf{x})^T \mathbf{\Lambda} \mathbf{\Phi} (\mathbf{x}')$ using the spectral distribution $\rho_{MP}^{\gamma} \boxtimes \rho_{MP}^{\psi}$ and sampling from independent Gaussian entries in lieu of the eigenfunctions (SUA). On the right, the simulated error is obtained by sampling from the prediction of kernel regression with the random (width-dependent) NNGP kernel $(\mathbf{x}, \mathbf{x}') \mapsto K^{\phi,N} (\mathbf{x}, \mathbf{x}')$ with $\phi = \mathrm{ReLU}$ nonlinearity on a subset of MNIST, and the theoretical error is obtained by sampling from the approximate random kernel $(\mathbf{x}, \mathbf{x}') \mapsto \mathbf{\Phi} (\mathbf{x})^T \mathbf{\Lambda} \mathbf{\Phi} (\mathbf{x}')$ using the spectral distribution of the actual NNGP kernel $\rho_{MP}^{\gamma} \boxtimes \mu_{\psi}^{\phi}$ and sampling from independent Gaussian entries in lieu of the eigenfunctions (SUA).}
\label{generrordistrib}
\end{figure}

To estimate the eigenfunctions $\Phi_i$, we rely on the spectral universality assumption (SUA), which is an important topic in the kernel literature \cite{Sollich_2002,El_Karoui_2010,doi:10.1142/S201032631350010X,Fan2015TheSN,kernelreghighdim,simon2022eigenlearning,lu2023equivalence}. The SUA states that in high dimension the eigenfunctions become unstructured and can be approximated by i.i.d.\ Gaussian entries $\mathbf{\Phi}_{i,j} \sim \mathcal{N} ( \mu, \sigma^2)$. In Figures \ref{kerneldistribisotrop} and \ref{kerneldistribrelumnist}, we illustrate to what extent this approximation and our equation \eqref{kernel identity} is correct on diagonal and off-diagonal elements of the random kernel matrix. This allows us to plot the distribution of generalization errors of the corresponding kernel regressions (Figure \ref{generrordistrib}). As expected, the spectral universality assumption provides a close match for (width-dependent) NNGP kernel with isotropic data and no nonlinearity, for which it has been proven to be exactly correct in infinite dimensions \cite{El_Karoui_2010}. The new insight here is to combine the SUA with the spectral distribution $\rho_{MP}^{\gamma} \boxtimes \mu_{\psi}^{\phi}$. On the other hand, there are some discrepancies in the case of MNIST with ReLU nonlinearity but the overall agreement in terms of generalization error is acceptable (Figure \ref{generrordistrib}). This observation is consistent with that of \cite{simon2022eigenlearning}. We then used these generalization error estimates for varying values of $\gamma = \frac{n}{N}$ to construct the double-descent curves (Figure \ref{generrorfigsynth}). Note however, that although without the SUA, we would have no way to estimate coefficients $A(\gamma,\psi), \bar{A} (\gamma,\psi), B (\gamma,\psi), C (\gamma,\psi), D (\gamma,\psi), \bar{D} (\gamma,\psi)$, as long as they are bounded and $B (\gamma,\psi)$ is nonzero, our theoretical result on the double-descent phenomenon remains valid.

\end{document}